\documentclass[acmsmall,screen]{acmart}
\AtBeginDocument{%
  }

\usepackage{algorithm}
\usepackage{algorithmic}
\usepackage{enumerate}
\usepackage{booktabs}
\usepackage{multirow}
\usepackage{xcolor}
\usepackage{caption}
\usepackage{enumitem}
\usepackage{wrapfig}
\usepackage{xspace}

\newcommand{\sysname}{\texttt{InfraMind}\xspace}

\renewcommand\footnotetextcopyrightpermission[1]{} 
\acmDOI{}
\acmISBN{}
\acmConference{}{}{}
\acmYear{}
\acmPrice{}

\begin{document}

\title{\sysname: A Novel Exploration-based GUI Agentic Framework for Mission-critical Industrial Management}


\author{Liangtao Lin}
\affiliation{%
  \institution{Nanyang Technological University}
  \country{Singapore}
}
\email{liangtao.lin@ntu.edu.sg}

\author{Zhaomeng Zhu}
\affiliation{%
  \institution{Nanyang Technological University}
  \country{Singapore}
}
\email{zhaomeng.zhu@ntu.edu.sg}

\author{Tianwei Zhang}
\author{Yonggang Wen}
\affiliation{%
  \institution{Nanyang Technological University}
  \country{Singapore}
}
\email{tianwei.zhang@ntu.edu.sg}
\email{ygwen@ntu.edu.sg}

\begin{abstract}
  Mission-critical industrial infrastructure, such as data centers, increasingly depends on complex management software. Its operations, however, pose significant challenges due to the escalating system complexity, multi-vendor integration, and a shortage of expert operators. While Robotic Process Automation (RPA) offers partial automation through handcrafted scripts, it suffers from limited flexibility and high maintenance costs. Recent advances in Large Language Model (LLM)-based graphical user interface (GUI) agents have enabled more flexible automation, yet these general-purpose agents face five critical challenges when applied to industrial management, including unfamiliar element understanding, precision and efficiency, state localization, deployment constraints, and safety requirements.
  To address these issues, we propose \sysname, a novel exploration-based GUI agentic framework specifically tailored for industrial management systems. \sysname integrates five innovative modules to systematically resolve different challenges in industrial management: (1) systematic search-based exploration with virtual machine snapshots for autonomous understanding of complex GUIs; (2) memory-driven planning to ensure high-precision and efficient task execution; (3) advanced state identification for robust localization in hierarchical interfaces; (4) structured knowledge distillation for efficient deployment with lightweight models; and (5) comprehensive, multi-layered safety mechanisms to safeguard sensitive operations.
  Extensive experiments on both open-source and commercial DCIM platforms demonstrate that our approach consistently outperforms existing frameworks in terms of task success rate and operational efficiency, providing a rigorous and scalable solution for industrial management automation.
\end{abstract}






\maketitle
\section{Introduction}
Mission-critical industrial management systems, including those used in electric grids, water treatment plants, railway systems, and data centers, serve as the digital command centers of modern infrastructure \cite{YADAV2021100433MCI4,huang2021dataDCIM}. Their provide operators with real-time monitoring, coordinated control, and intelligent decision support, thus meeting the ever-growing and increasingly complex management demands of critical infrastructure \cite{en18133576newMCI2,1c89305696a346e386c44c89c8fff229dds}. Ensuring the safe, efficient, and reliable operation of these systems is essential for public safety and economic stability \cite{mcdaniels2007empiricalMCI3, YADAV2021100433MCI4}.

However, as industrial digitalization accelerates and systems become increasingly interconnected, mission-critical management platforms face three persistent operational challenges: 
(i) the escalating complexity and customization of industrial management software, which increasingly integrates diverse functions and interfaces from multiple vendors \cite{siemens1-2021claritylc, schneider3-2025_complexitiesGridMod};
(ii) the necessity for coordinated, real-time management across multiple, interdependent subsystems (e.g., power, cooling, IT, security) \cite{8783313crosssystem, greengrid_dcim}; and
(iii) a pronounced scarcity of experienced operators, with high rates of operational errors frequently reported in practice \cite{uptime7-2023StaffingSurvey,uptime8-2019_tryharder}.
These realities create an urgent need for reliable, intelligent automation solutions to support safe and efficient management at scale \cite{zangana2025leveragingMCI6}. 

One of the most widely adopted approaches to industry automation is Robotic Process Automation (RPA), which aims to address the growing demand for reliable and efficient operations \cite{lamberton2017impactRPA8,9001110RPA9}. RPA typically automates workflows by scripting sequences of predefined user interactions, such as mouse clicks, keystrokes, and data entry, thus handling repetitive and rule-based tasks within industrial GUIs. This approach can significantly reduce manual workload and minimize human error for routine operations. However, RPA solutions are heavily reliant on handcrafted scripts and static rules \cite{chawla2024guideRPAweak}, requiring substantial manual maintenance whenever interfaces change or new workflows are introduced. Such rigidity limits RPA’s suitability for complex, dynamic, or evolving industrial systems, where adaptability and scalability are essential.

The advent of Large Language Models (LLMs) opens up a new paradigm for automating mission-critical industrial management systems through intelligent GUI agents. Compared to RPA, LLM-based agents \cite{hu-etal-2025-osagents,qin2025uitars} are capable of interpreting unstructured interfaces, reasoning about complex tasks, and executing flexible, context-aware operations. 
However, several critical challenges remain for applying existing GUI agents in industrial domains.
(1) \textbf{Unfamiliar Elements:} they are trained on general web or consumer software, thus failing to infer the functions of complex and unfamiliar interface elements.
(2) \textbf{Precision \& Efficiency:} they agents are incapable of dynamic trial-and-error exploration to meet stringent requirements for precision and efficiency.
(3) \textbf{State Localization:} there is a lack of effective mechanisms for state identification and hierarchical localization in desktop-based industrial GUIs without explicit indicators.
(4) \textbf{Deployment Constraints:} they do not have sufficient consideration of deployment constraints in network-isolated or resource-limited environments.
(5) \textbf{Safety Requirements:} they do not have built-in safety mechanisms for managing sensitive or safety-critical operations.


To overcome these limitations, we propose \textit{\sysname}, the first exploration-based agentic framework tailored for mission-critical industrial management. \sysname autonomously learns the functional structures and operational workflows of complex industrial GUIs, enabling efficient and reliable automation across diverse, customized interfaces. By integrating systematic exploration mechanisms and powerful modules, \sysname addresses the core challenges of adaptability, knowledge acquisition, and safe deployment in sensitive operational contexts, providing a practical and scalable solution for automating industrial management systems.

Specifically, \sysname adopts an exploration-centric paradigm, structured around a centralized multi-agent architecture and a Perception–Reasoning–Action loop. It incorporates the following innovative modules to address the aforementioned challenges: (1) \textbf{Systematic Exploration} enables autonomous discovery and semantic grounding of complex and unfamiliar GUI elements through search-based exploration and icon–caption knowledge base construction; (2) \textbf{Memory-Driven Planning} facilitates high-precision and efficient task execution by summarizing, and reusing optimal action-flow plans derived from prior task exploration; (3) \textbf{Advanced State Identification} achieves robust localization and workflow tracking within hierarchical desktop GUIs by integrating semantic–visual representations and constructing a state transition graph; (4) \textbf{Knowledge Distillation} transfers operational knowledge from large models to compact models, enabling efficient and fully offline deployment under industrial resource constraints; (5) \textbf{Comprehensive Safety Mechanisms} ensure reliable operation in safety-critical environments through multi-layered safeguards, including blacklist filtering, operator confirmation, and LLM-based risk assessment.


To rigorously evaluate our framework, we constructed a benchmark based on two industrial-grade Data Center Infrastructure Management (DCIM) platforms, representing typical mission-critical infrastructure scenarios. Our \sysname demonstrated excellent ability to efficiently learn, adapt, and automate complex operational workflows across both systems. Extensive evaluations revealed that our framework consistently achieved higher task success rates and greater efficiency than state-of-the-art GUI Agents, including OmniTool ~\cite{lu2024omniparserpurevisionbased}, Agent S2 ~\cite{Agent-S2} and UI-TARS-1.5 ~\cite{qin2025uitars}. Ablation studies confirmed the critical role of systematic exploration and planning modules, and further demonstrated that operational knowledge acquired by large vision-language models (VLMs) can be effectively transferred into smaller, deployment-ready models. Finally, case studies and real-world data center deployments validate the effectiveness of our exploration-based approach, supporting robust automation that streamlines task execution and reduces operator cognitive load. These results collectively highlight the reliability, practical value, and broad deployment potential of our framework across diverse mission-critical environments.

Our major contributions include:
\begin{itemize}[noitemsep,leftmargin=*]
    \item We introduce \sysname, the first exploration-centric agentic framework tailored for mission-critical industrial software. It leverages state-of-the-art VLMs to achieve robust operational automation. By utilizing VM snapshots and systematic exploration, our framework overcomes the fundamental challenge of irreversibility in GUI agent learning, allowing safe and efficient acquisition of operational knowledge, mastering complex and unfamiliar software elements, and achieving accurate state localization within industrial management GUIs.
    
    \item We propose a unified methodology for GUI agent knowledge representation, encompassing three complementary structures: Icon–caption pairs, planning trees, and state transition graphs. These are acquired by large models and transferred to lightweight models, allowing compact models such as Qwen2.5-VL-7B to achieve comparable performance with GPT-4o in industrial automation, thus supporting efficient deployment under practical resource constraints.
    
    \item To address the safety-critical nature of industrial automation, our framework integrates a multi-layered safety module with proactive risk controls throughout both exploration and execution. This includes CLIP-based blacklisting of risky GUI elements, operator-in-the-loop hazard confirmation, and LLM-driven semantic risk assessment, together ensuring reliable, transparent, and secure agent operation in sensitive and high-stakes scenarios.
    
    \item We conduct extensive experiments on both open-source and commercial DCIM platforms, using data center as a representative scenario. Our results demonstrate significant improvements in task success rates and operational efficiency compared to existing frameworks, addressing the stringent precision and efficiency requirements of industrial automation, and providing a robust reference solution for deploying GUI agents across a wide range of critical industrial domains.
    
\end{itemize}

\section{Background and Motivation}

\subsection{Mission-critical Industrial infrastructure and Management}
Mission-critical infrastructure, including electric grids, water treatment plants, railway systems, and data centers, forms the indispensable technological backbone of modern society. These complex systems underpin essential services that must operate continuously and reliably to ensure public safety, economic stability, and seamless functioning of daily life \cite{de2007systemsMCI1,JOHANSSON201327MCI2}. Even brief disruptions or failures in these environments can cause widespread societal and financial repercussions, from large-scale service outages to severe security and safety risks \cite{mcdaniels2007empiricalMCI3, YADAV2021100433MCI4}. Accordingly, the operational efficiency, reliability, and resilience of such infrastructure are paramount, driving ongoing investment in advanced management and automation technologies \cite{s25061666MCI5,zangana2025leveragingMCI6,houliotis2018missionMCI7}.

At the core of modern mission-critical infrastructure lies a diverse array of industrial management software platforms, also known as industrial control or management systems. These software solutions are responsible for a range of essential functions, including real-time system monitoring, coordinated process control, event and alarm management, asset and resource tracking, data analytics, workflow automation, and remote operation of complex equipment. Operators depend on these platforms for situational awareness, intelligent decision support, rapid fault detection, and prompt intervention, often across highly heterogeneous and interconnected subsystems.

The demands on management software are significant: systems must provide accurate, real-time visibility into operational status, support seamless integration with multi-vendor hardware, enable timely and coordinated responses to incidents, and ensure continuous compliance with rigorous safety and reliability standards. As the scale and complexity of critical infrastructure continues to grow, driven by digitalization, multi-system integration, and rising performance expectations, management software is expected to support not only high efficiency and operational reliability, but also rapid adaptation and robust automation in dynamic, resource-constrained environments.

A prominent example is Data Center Infrastructure Management (DCIM) platforms, which serve as the digital command centre of modern data centers. DCIM software integrates monitoring and control across critical subsystems, such as power, cooling, IT equipment, and security, enabling operators to track equipment status, manage resources, visualize operational data, automate workflows, and respond rapidly to anomalies. These platforms are characterized by the need for high availability, seamless interoperability with diverse hardware, and strong automation to ensure safe, efficient, and resilient data center operations.

In this work, we select DCIM as a representative scenario of mission-critical industrial management. Our study focuses on the unique operational challenges, functional requirements, and automation needs present in modern data center environments, using DCIM as the primary context for evaluating and advancing intelligent management solutions.

\subsection{Challenges of Management Automation}
\label{sec:challenge}
Recent advances in LLM-based agents have greatly accelerated the development of GUI automation. However, mission-critical industrial management systems present a set of unique challenges, which cannot be addressed by existing GUI agent frameworks. These challenges are particularly acute in industrial domains, where the stakes of automation are high and the constraints are fundamentally different from those of typical web or consumer applications.

\begin{figure*}[t]
    \centering
    \includegraphics[width=\linewidth]{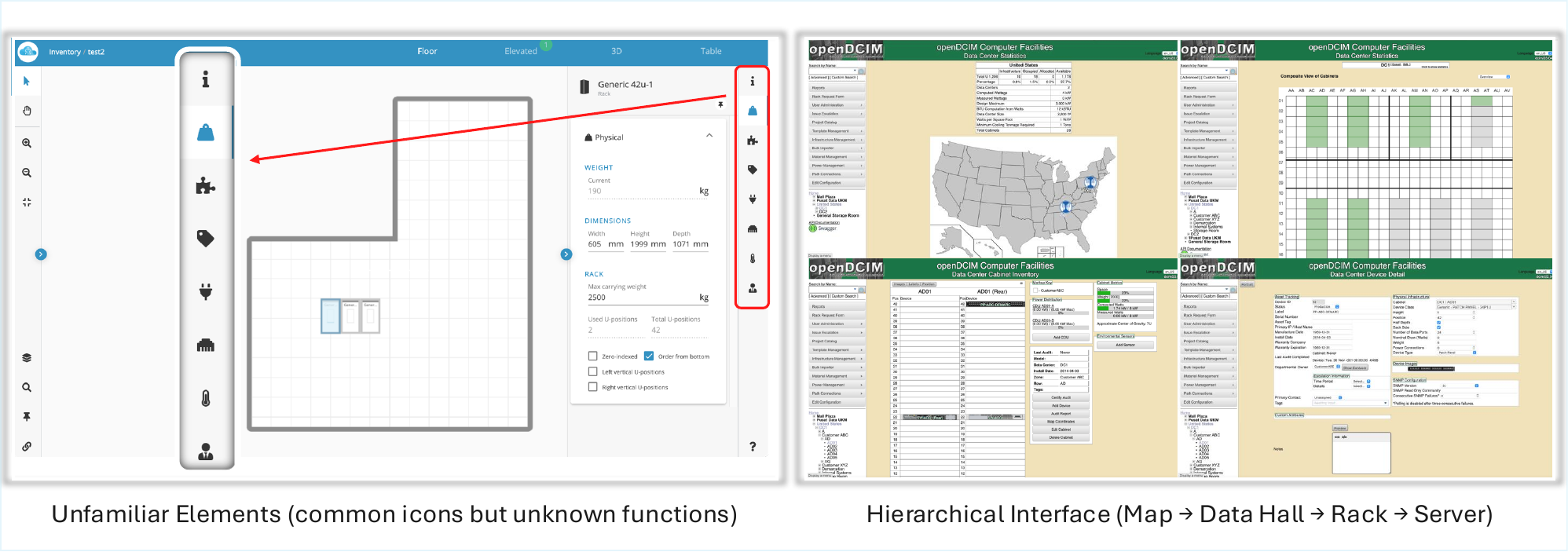}
    \caption{\textbf{Sample GUI Challenges in Industrial Management Systems.} C1: Unfamiliar Interface Elements (Left); C3: Hierarchical Interface Structures (Right).}
    \label{fig:f0}
\end{figure*}

\noindent \textbf{$\bullet$ C1: Complex and Unfamiliar Interface Elements.} 
Industrial management GUIs frequently incorporate highly specialized or custom-developed controls and panels.
These uniquely labeled or icon-based elements cannot be interpreted by general-purpose agents, as they are typically trained on web or office environments. This poses persistent barriers to accurate interface understanding and robust task execution.
For example, in Schneider EcoStruxure IT \cite{schneider_ecostruxure_it} (see Figure~\ref{fig:f0}, left), the sidebar features a range of symbols for device overviews, physical layouts, and usage statistics. However, these buttons often use generic or proprietary icons whose actual functions are not apparent from appearance alone, making it difficult even for human operators to discern their purpose without direct interaction.

\noindent \textbf{$\bullet$ C2: Stringent Requirements for Precision and Efficiency.} 
Industrial environments demand extremely high precision and operational efficiency during automation. However, current GUI agent frameworks often rely on real-time, unstable exploration without persistent memory, resulting in inconsistent outcomes and inefficiencies that are unacceptable in mission-critical settings.  
For instance, in data center operations, human operators frequently interact with DCIM platforms to quickly locate and monitor specific equipment, performing rapid adjustments in response to dynamic operational demands. Agents that fail to deliver stable and efficient navigation can introduce unacceptable delays or errors in such high-stakes contexts.

\noindent \textbf{$\bullet$ C3: State Identification and Hierarchical Localization.} 
Industrial management GUIs are typically implemented as standalone desktop applications rather than web-based systems, and thus lack explicit state identifiers such as URLs. This makes it exceptionally difficult for GUI agents to recognize and track their precise position or progress within complex, hierarchical interface structures. As a result, agents often cannot determine where they are in the workflow or which step has been completed, leading to significant challenges in planning subsequent actions or recovering from errors.  
A typical example can be found in DCIM (see Figure~\ref{fig:f0}, right), where interfaces are organized as asset trees (e.g., Data Hall $\rightarrow$ Rack $\rightarrow$ Server). Without standard addressable indicators, agents must attempt to infer their current state in the hierarchy, making it challenging to perform robust reasoning, effective task planning, and reliable error recovery.

\noindent \textbf{$\bullet$ C4: Offline Operation and Deployment Constraints.} 
Many industrial management systems operate in network-isolated or air-gapped environments, requiring GUI agents to function fully offline and independent of cloud-based resources. This poses significant challenges for model deployment, updating, and resource management that are seldom considered by existing frameworks.  
While some recent works target mobile devices with lightweight agents, such approaches are rarely evaluated in complex, offline industrial environments like data centers, where both capability and efficient deployment must be simultaneously achieved.

\noindent \textbf{$\bullet$ C5: Safety-Critical and Sensitive Operations.} 
Industrial GUIs are characterized by the presence of safety-critical controls, such as emergency shutdown, reset, or override functions. Erroneous actions can lead to severe operational consequences, demanding sophisticated risk awareness and proactive safety mechanisms far beyond what current agents offer.  
For example, DCIM platforms often provide direct controls for server addition or removal, power toggling, and system configuration. Mistaken or unsafe actions in these contexts may result in equipment damage, service interruptions, or data loss, underscoring the necessity of fine-grained, context-aware safety modules.

Addressing these domain-specific challenges is essential for the practical deployment of GUI agents in industrial management systems. Existing automation paradigms and agent frameworks are not directly adopted, underscoring the necessity and novelty of our proposed approach.

\section{Methodology}

\begin{figure*}[t]
    \centering
    \includegraphics[width=\linewidth]{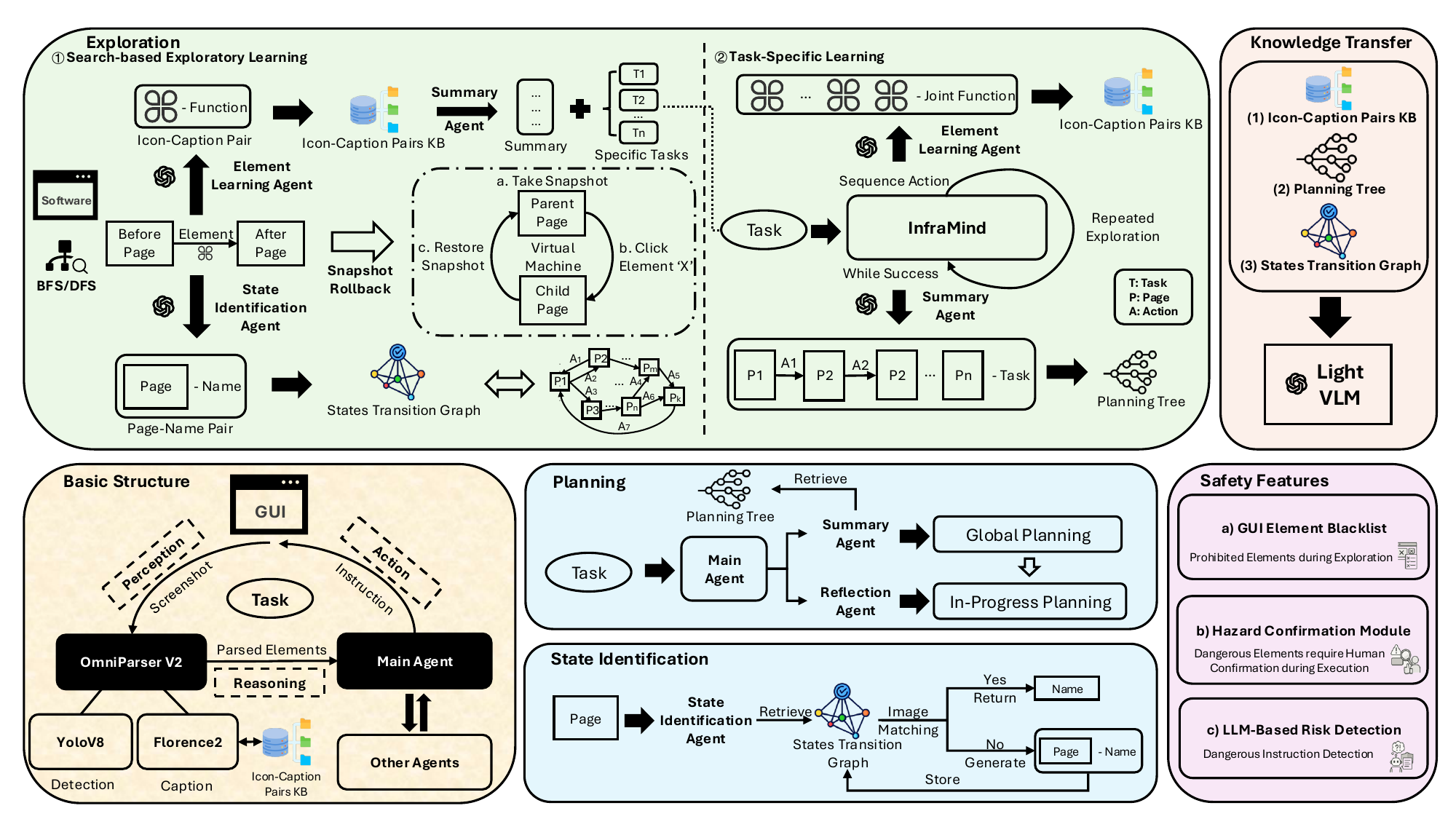}
    \caption{\textbf{Architecture of our \sysname framework.}}
    \label{fig:f1}
\end{figure*}

We design \sysname, a novel modular agentic workflow for robust and adaptive automation of complex industrial GUIs. As illustrated in Figure~\ref{fig:f1}, \sysname is a centralized multi-agent architecture, following the Perception–Reasoning–Action paradigm. At its core, a centralized \emph{Main Agent} coordinates the overall workflow and delegates tasks to several specialized agents: a \emph{Summary Agent} (responsible for aggregating successful action flows, retrieving prior experience, and generating software capability summaries), a \emph{Reflection Agent} (which dynamically adjusts in-progress plans during execution), an \emph{Element Learning Agent} (dedicated to learning and updating knowledge of GUI elements), and a \emph{State Identification Agent} (which handles all aspects of state recognition and localization).

To address the challenges in Section \ref{sec:challenge}, \sysname incorporates multiple innovative modules, including systematic exploration (C1), memory-driven planning (C2), advanced state identification (C3), knowledge distillation (C4) and comprehensive safety mechanisms (C5). Below we elaborate the design of those modules and how they resolve the challenges. 



\subsection{Understanding Complex and Unfamiliar Interface Elements (C1)}
\label{sec:systematic-exploration}
A fundamental obstacle in automating industrial management GUIs is the presence of numerous domain-specific, custom-developed elements whose functions are not readily understood by general-purpose GUI agents. Overcoming this challenge requires a principled approach to grounding, that is, enabling the agent to correctly map unfamiliar GUI elements to their semantic functions.


Currently, GUI agent methods typically follow one of three paradigms: 

\begin{itemize}[noitemsep,leftmargin=*]
    \item \textbf{DOM-tree-based parsing}, which directly extracts interactive elements from webpage source code \cite{browser_use2024}, but is limited to web interfaces and ineffective for custom industrial software elements. 
    
    \item \textbf{Detection-based parsing}, which initially detect interactive elements through object detection models or similar techniques \cite{singh2025trishul,lu2024omniparserpurevisionbased}, subsequently passing the identified elements and coordinates to VLMs, thus providing higher adaptability to arbitrary visual environments.
    
    \item \textbf{Direct grounding via VLM}, which leverage the grounding capabilities of VLMs to simultaneously reason and specify interaction coordinates \cite{qin2025uitars,Agent-S,Agent-S2}. However, these methods require highly capable VLMs that are typically trained on generic data, and thus may struggle to generalize to specialized elements in industrial software.
\end{itemize}

As discussed above, the first paradigm is fundamentally inapplicable to industrial desktop GUIs, while the third is constrained by the scarcity of specialized training data for professional domains. The detection-based approach, by contrast, offers the best balance between adaptability and extensibility: it allows for the identification of arbitrary on-screen elements but still faces the crucial bottleneck of accurately captioning their functions.

\subsubsection{Learning Icon–Caption Pairs through Systematic Search-based Exploration}
To address this bottleneck, we propose a novel, learning-driven grounding mechanism: rather than relying on prior semantic knowledge, the agent systematically acquires icon–caption pairs specific to the target industrial management software through autonomous exploration.

Specifically, as shown in Algorithms~\ref{alg:bfs-explore} and \ref{alg:dfs-explore}, \sysname operates within a virtual machine environment, employing  Breadth-First Search (BFS)/ Depth-First Search (DFS) search strategies combined with VM snapshot and rollback capabilities. At each GUI state, \sysname interacts with every detected element, observes the resulting visual and semantic changes, and records pre- and post-action screenshots. An \textit{Element Learning Agent} is then prompted to summarize the function of each element by comparing the interface before and after the interaction, generating candidate captions even for previously unseen elements.

To the best of our knowledge, we are the first to systematically introduce VM-based snapshot and rollback to address the fundamental challenge of irreversibility and non-idempotent in GUI agent exploration, which is a core issue across the broader field of GUI agent research. By encapsulating every exploratory action within a retrievable VM state, the agent can safely and exhaustively explore the full action space. This reversible exploration mechanism not only ensures robust knowledge acquisition and operational safety, but also establishes a solid technical foundation for future reinforcement learning by GUI agents.\\

\begin{algorithm}[t]
\caption{BFS-Based GUI Exploration with VLM-based Function Summarization and VM Snapshot Rollback}
\label{alg:bfs-explore}
\begin{algorithmic}[1]
\REQUIRE Initial GUI state $s_0$
\STATE Initialize a queue $Q \leftarrow [ s_0 ]$
\STATE Initialize a set of visited states $\mathcal{V} \leftarrow \{\}$
\WHILE{$Q$ is not empty}
    \STATE Pop the front state $s$ from $Q$
    \IF{$s \notin \mathcal{V}$}
        \STATE Add $s$ to $\mathcal{V}$
        \STATE Take and save a VM snapshot for $s$ using $R_\mathrm{VM}(s)$
        \FOR{each clickable GUI element $e$ in $s$}
            \STATE Execute action $a_e$ (click $e$), observe new state $s'$
            \STATE Use $F_\mathrm{VLM}(s, a_e, s')$ to summarize the function of element $e$ based on the state change
            \STATE Record transition $(s, a_e, s', F_\mathrm{VLM}(s, a_e, s'))$
            \IF{$s' \notin \mathcal{V}$}  
                \STATE Append $s'$ to $Q$
            \ENDIF
            \STATE Restore the VM to snapshot for $s$ using $R_\mathrm{VM}^{-1}(s)$ (rollback)
        \ENDFOR
    \ENDIF
\ENDWHILE
\end{algorithmic}
\end{algorithm}

\begin{algorithm}[ht]
\caption{DFS-Based GUI Exploration with VLM-based Function Summarization and VM Snapshot Rollback}
\label{alg:dfs-explore}
\begin{algorithmic}[1]
\REQUIRE
    Initial GUI state $s_0$; \\
    A virtual machine (VM) supporting snapshot $R_\mathrm{VM}(s)$ and rollback $R_\mathrm{VM}^{-1}(s)$; \\
    A vision-language model (VLM) function $F_\mathrm{VLM}(s, a_e, s')$
\STATE Initialize a stack $S \leftarrow [s_0]$
\STATE Initialize a set of visited states $\mathcal{V} \leftarrow \{\}$
\WHILE{$S$ is not empty}
    \STATE Pop the top state $s$ from $S$
    \IF{$s \notin \mathcal{V}$}
        \STATE Add $s$ to $\mathcal{V}$
        \STATE Take and save a VM snapshot for $s$ using $R_\mathrm{VM}(s)$
        \FOR{each clickable GUI element $e$ in $s$}
            \STATE Execute action $a_e$ (click $e$), observe new state $s'$
            \STATE Use $F_\mathrm{VLM}(s, a_e, s')$ to summarize the function of element $e$ based on the state change
            \IF{$s' \notin \mathcal{V}$}
                \STATE Record transition $(s, a_e, s', F_\mathrm{VLM}(s, a_e, s'))$
                \STATE Push $s'$ onto $S$
            \ENDIF
            \STATE Restore the VM to snapshot for $s$ using $R_\mathrm{VM}^{-1}(s)$ (rollback)
        \ENDFOR
    \ENDIF
\ENDWHILE
\end{algorithmic}
\end{algorithm}

\subsubsection{Constructing and Enhancing the Icon–Caption Knowledge Base}
The results of this systematic exploration are aggregated to form a dedicated icon–caption knowledge base, which is then leveraged via a dual, knowledge-guided strategy to enhance caption quality and robustness:

\begin{itemize}[noitemsep,leftmargin=*]
    \item \textbf{VLM fine-tuning:} The collected icon–caption pairs are used to fine-tune the VLM (Florence2 in OmniParser V2), allowing it to generate accurate, domain-specific captions for GUI elements.
    \item \textbf{CLIP-based image matching:} At inference time, the knowledge base supports robust caption retrieval by matching detected icons against the database using CLIP-based similarity \cite{radford2021learningclip}, ensuring consistent and contextually appropriate annotations even as interfaces evolve.
\end{itemize}
This dual strategy enables both generative and retrieval-based captioning, yielding structured GUI representations annotated with high-quality semantic labels.

This learning-driven approach empowers \sysname to operate effectively in industrial GUIs that deviate from standard designs, equipping it with domain-adapted semantic understanding that general-purpose agents lack.

\subsection{Achieving High Precision and Efficiency through Prior Knowledge (C2)}
Industrial management tasks require automation agents not only to be robust and adaptive, but also to execute tasks with high precision and operational efficiency. This is particularly critical in real-world settings where operational delays, repeated failures, or trial-and-error actions are unacceptable. To this end, we focus on prior knowledge acquisition, enabling the agent to learn, store and reuse optimal execution plans before actual deployment. It consists of the following steps. 

\subsubsection{Software Abstraction and Task Preprocessing}
Before \sysname can effectively accumulate actionable procedural knowledge, we first perform data preprocessing based on the results of the previous systematic GUI exploration step (Section \ref{sec:systematic-exploration}). Here, a \textit{Summary Agent} is tasked to synthesize a high-level summary of the software’s interface structure and available functional regions. Based on this summary, the agent automatically generates a set of representative and executable tasks that reflect the core workflows of the system. This preparatory step frames the agent’s subsequent experience collection.

\begin{figure}
    \centering
    \includegraphics[width=0.5\linewidth]{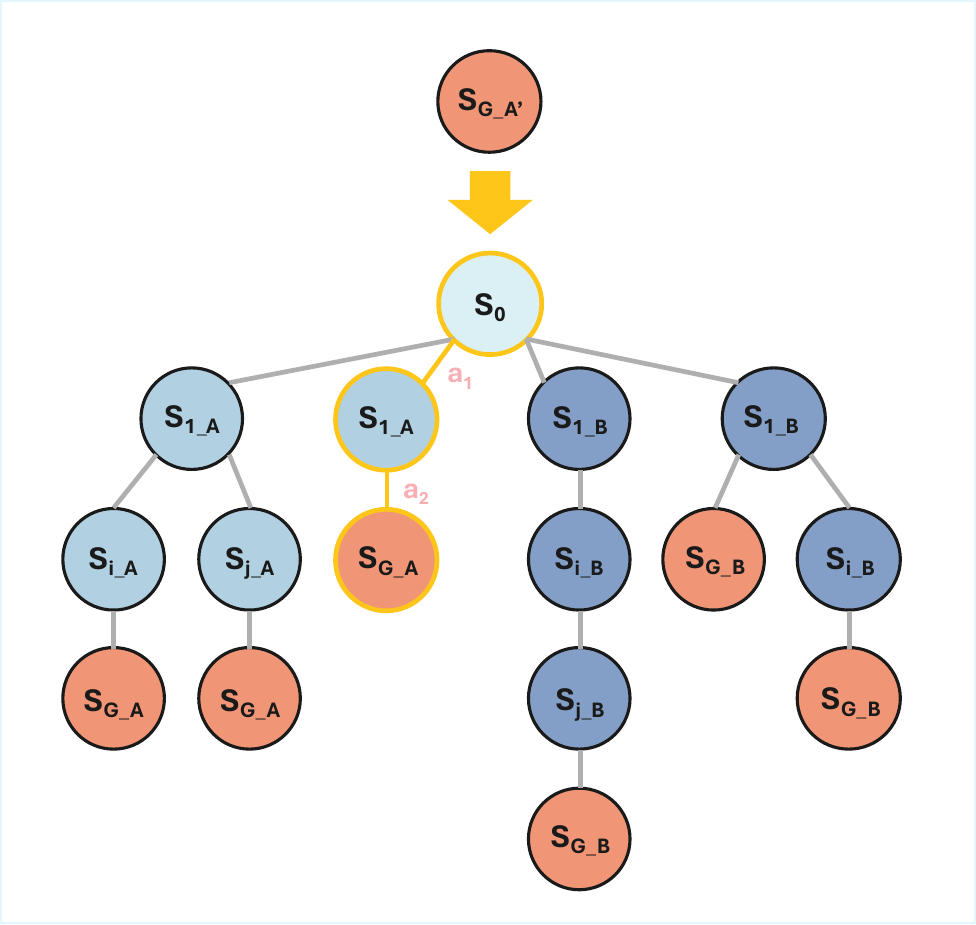}
    \vspace{-10pt}
    \caption{\textbf{Example of the action-flow tree}}
    \vspace{-10pt}
    \label{fig:f3}
\end{figure}
\subsubsection{Task-Specific Learning and Action-Flow Construction} 
Building upon the task abstraction, \sysname enters another core learning stage: it autonomously attempts each representative task in the virtual environment, systematically exploring different action sequences to identify successful solutions. For every success, the \textit{Summary Agent} captures the sequence of GUI states and user actions as an \textit{action-flow tree}, a structured representation in which nodes correspond to interface states and edges denote actions. 

Figure \ref{fig:f3} gives an example of an action-flow tree \( \mathcal{T}_{s_G} \), which aggregates successful trajectories from multiple historical tasks. Each node represents a GUI-observable state \( s_t \in \mathcal{S} \), and each edge denotes a GUI action \( a_t \in \mathcal{A} \). The root node \( s_0 \) corresponds to the shared initial screen (typically the system homepage), while leaf nodes in orange indicate distinct goal states \( s_G \) from previous tasks. The tree is partitioned into different colour-coded subtrees, each corresponding to a different historical task domain. 

Through repeated execution and incremental refinement, \sysname accumulates a comprehensive library of successful task flows, thereby transforming empirical trial-and-error into reusable procedural knowledge.

\subsubsection{Experience-Guided Planning and Task Execution}
During real-world deployment, the Summary Agent leverages this action-flow tree to guide new executions. As shown in Figure~\ref{fig:f3}, when presented with a task, it retrieves the most relevant subtree corresponding to prior successful experiences, identifies the shortest successful trajectory to the target, and generates a global plan for \sysname. This experience-guided retrieval not only improves task success probability but also greatly enhances efficiency, as \sysname can reuse optimal action sequences learned offline.

\subsubsection{Additional Learning of Compound Actions}
Throughout this process, the Element Learning Agent also observes and internalizes the behavior of compound or interdependent elements, such as multi-step forms, confirmation dialogs, and context-dependent controls, enriching its repertoire and further increasing automation reliability.

In summary, by transforming one-off, trial-and-error attempts into a structured, reusable repository of action-flows, our framework enables reliable, efficient, and knowledge-driven automation, in direct response to the stringent precision and efficiency requirements of industrial applications.

\subsection{Robust State Identification and Localization (C3)}
To enable robust task automation in industrial desktop GUIs, our agent must accurately identify and localize its current state, even in the absence of explicit identifiers such as URLs. We therefore introduce a dedicated \textit{State Identification Agent} that integrates both semantic and visual cues to achieve precise localization throughout the automation workflow.

\subsubsection{Semantic–Visual State Representation}
During the first exploration phase, for each encountered interface, the State Identification Agent generates a comprehensive textual description, summarizing layout structure, salient features, and functional regions. Simultaneously, the agent applies a CLIP-based image similarity model to encode visual state representations. By combining these semantic and visual embeddings, the system can reliably distinguish between highly similar or dynamic interface states. As \sysname explores possible interactions, each unique interface state is indexed and organized into a directed state transition graph.



\subsubsection{Reliable State Localization and Planning}

With this graph-based abstraction, the agent can not only localize its current position but also resume interrupted workflows, perform error recovery, and plan structured navigation to target states. The unified state graph further allows the agent to generalize across repetitive or hierarchical UI patterns, significantly enhancing scalability and adaptability to heterogeneous industrial software environments.

By tightly coupling semantic analysis, visual matching, and graph-based modeling, our State Identification Agent overcomes the ambiguity and complexity inherent to industrial GUIs, forming the backbone of reliable, context-aware automation.

\subsection{Efficient Deployment via Knowledge Transfer (C4)}

A major obstacle to deploying advanced GUI agents in industrial environments is the requirement for fully offline operation and efficient resource usage. Many mission-critical systems are network-isolated or air-gapped, making it impossible to rely on cloud-based APIs such as GPT-4o or Gemini. Even when local deployment is permitted, large-scale models such as Qwen2.5-VL-32B, despite their strong reasoning and perception abilities, are often too computationally demanding for real-time use in industrial environments due to their high latency and resource consumption.

In contrast, a smaller model such as Qwen2.5-VL-7B is much more suitable for on-premise deployment, but may lack the advanced capabilities required for robust GUI automation out-of-the-box. Some recent research has explored transferring the capabilities of large models to smaller ones via knowledge distillation or structured artifact transfer~\cite{jang2025vl2lite,10.1145/3699518knowdistsurvey}. Inspired by this, we propose a knowledge transfer framework tailored for industrial GUI automation.

\subsubsection{Structured Knowledge Transfer for Industrial Deployment}
During the exploration and learning phase, all heavy perception, reasoning, and planning tasks are performed using the large-scale model. Through this process, \sysname constructs three key structured knowledge bases:

\begin{itemize}[noitemsep,leftmargin=*]
    \item \textbf{GUI Element Functionalities:}  
        Icon–caption pairs mapping interface elements to their learned semantic functions.
    \item \textbf{Execution Plans:}  
        Action-flow trees representing successful task trajectories across varied workflows.
    \item \textbf{Interface State Transitions:}  
        Labeled interface states indexed by visual similarity and structured as a state transition graph.
\end{itemize}

\subsubsection{Efficient Deployment with Lightweight Models} 
At deployment time, only the compact model is used, running fully offline and consulting these knowledge repositories to ground interface elements, plan actions, and track its state. This design enables \sysname to achieve practical performance close to that of large models, while remaining lightweight enough for mission-critical, resource-constrained industrial environments.

In summary, our knowledge transfer strategy effectively bridges the gap between the capability of large-scale models and the deployment constraints of industrial automation, allowing small models to deliver reliable and efficient automation in settings where cloud APIs and high-end hardware are infeasible.

\subsection{Ensuring Safety in Sensitive Operations (C5)}
Given the sensitive nature of mission-critical industrial software, safety is a central concern in our framework. 
GUI automation inevitably involves interaction with potentially hazardous elements. The most direct solution is to prohibit operations on these elements during both exploration and execution. Accordingly, our framework actively detects and blocks high-risk elements throughout, incorporates operator confirmation for critical actions, and further enhances safety through semantic-level risk assessment of planned operations.

Specifically, we implement a multi-layered set of built-in safety mechanisms to proactively minimise operational risks during all stages of automation:

\subsubsection{GUI Element Blacklist}

During the BFS/DFS-based exploration phase, the system maintains a curated blacklist of prohibited UI elements and regions known to trigger dangerous or irreversible actions. For every interface, all detected elements are compared to blacklist entries using a CLIP-based visual matching approach, ensuring that any element resembling a blacklisted item is automatically excluded from exploration candidates. This mechanism prevents the agent from inadvertently interacting with high-risk areas during both automated learning and execution.

\subsubsection{Hazard Confirmation Module}  

When \sysname is about to perform an action on a potentially hazardous element, it cross-references the target with a database of critical or sensitive UI components using image matching. If a match is found, the system triggers a confirmation process: a pop-up dialog is displayed, prompting the user to review and explicitly approve or reject the action. Optionally, control can be handed over entirely to the human operator. This human-in-the-loop safeguard acts as a final barrier against unintended critical operations.

\subsubsection{LLM-Based Risk Detection}  

Prior to executing any planned instruction, \sysname leverages an LLM-as-Judge semantic risk assessment \cite{brown2024self-eval}. Given the user input or generated action sequence, the LLM evaluates whether the instruction is potentially harmful, hazardous, or outside safe operating bounds. If a risk is detected, the agent alerts the user and seeks confirmation before proceeding, providing an additional semantic layer of safety particularly effective for detecting nuanced or context-dependent risks.

Collectively, these safety mechanisms ensure that the agent operates with caution and transparency, significantly reducing the likelihood of unsafe or irreversible actions within mission-critical environments.


\section{Evaluation}
To evaluate the performance and robustness of \sysname in real-world GUI-based infrastructure management, we organize this section around the following three research questions:
\begin{itemize}[noitemsep,leftmargin=*]
    \item \textbf{RQ1 (\S4.2):} \textit{How effective is our \sysname in completing complex GUI tasks across diverse platforms?}
    We benchmark our framework against state-of-the-art agents on two heterogeneous DCIM systems, assessing success rate and efficiency under step constraints.
    \item \textbf{RQ2 (\S4.3):} \textit{How do key components and model choices influence \sysname performance?}
    We conduct ablation studies to isolate the contribution of exploration and planning modules, and examine how different vision-language models affect final outcomes.
    \item \textbf{RQ3 (\S4.4):} \textit{How does \sysname leverage learned knowledge to improve operational efficiency and ensure safety in practice?}
    Through case studies, we demonstrate how interface-specific knowledge enhances real-time decision-making and reduces redundant actions, and reinforces operational safety through integrated risk controls..
\end{itemize}

\subsection{Experimental Setup}

To rigorously evaluate our agentic automation framework in mission-critical infrastructure scenarios, we curated a benchmark focused on data center operations. We selected two representative DCIM systems: the open-source platform \textit{OpenDCIM} \cite{opendcim}, and Schneider Electric’s \textit{EcoStruxure IT} \cite{schneider_ecostruxure_it}, a widely used commercial solution. For each platform, we manually designed 10 representative tasks, spanning three levels of difficulty (easy, medium, and hard), to comprehensively cover typical operational requirements encountered in real-world data centers. 

The difficulty of the tasks are defined as follows:
\begin{itemize}[noitemsep,leftmargin=*]
    \item \textbf{Easy Tasks}: Operations that require a single-step action or direct information retrieval within the current screen (e.g., checking for alerts, clicking clearly labeled entries).
    \item \textbf{Medium Tasks}: Tasks involving multi-step navigation across 2–3 interface layers before executing a straightforward operation (e.g., accessing a specific data center’s overview panel).
    \item \textbf{Hard Tasks}: More complex or composite workflows, often requiring deep GUI traversal, context-sensitive decision making, or multi-object operations (e.g., adding or deleting servers in a rack).
\end{itemize}

Each task was executed three times per agent to account for potential variability in performance. All experiments were conducted under a maximum step constraint of 20 steps per task: if an agent failed to complete a task within 20 steps, the run was marked as a failure and assigned 20 steps for averaging. Each agent was evaluated on all tasks, and both Success Rate and Average Steps to completion were reported as key metrics. Task success was verified through human evaluation, where annotators confirmed whether the final system state matched the expected outcome.

We benchmark our system against three leading agentic automation frameworks:

\begin{itemize}[noitemsep,leftmargin=*]
\item\textbf{OmniTool} \cite{lu2024omniparserpurevisionbased}:
A Microsoft-developed GUI agent toolkit that integrates the OmniParser engine with vision-language models. OmniTool enables flexible control of a Windows 11 VM through visual parsing and supports a wide range of models. In our evaluation, we select OmniParser V2 paired with the GPT-4o vision model as a representative.

\item\textbf{Agent S2} \cite{Agent-S2}:
A compositional framework that distributes tasks across generalist and specialist models. With novel grounding and hierarchical planning techniques, Agent S2 sets new benchmarks on GUI automation tasks. We choose Claude-3.7-Sonnet as its core model.

\item\textbf{UI-TARS-1.5} \cite{qin2025uitars}:
An open-source agent from ByteDance, excelling at screenshot-based GUI interaction across platforms. It features unified action modeling, advanced perception, and strong adaptability through reflective training, achieving SOTA performance on multiple benchmarks.
\end{itemize}

These frameworks represent the current state-of-the-art and provide robust baselines for evaluating our approach.

\begin{table*}[ht!]
\centering
\begin{tabular}{lcccc}
\toprule
\multirow{2}{*}{\textbf{Agentic Framwork}} & \multicolumn{2}{c}{\textbf{OpenDCIM}} & \multicolumn{2}{c}{\textbf{EcoStruxure IT}} \\
\cmidrule(lr){2-3} \cmidrule(lr){4-5}
 & Success Rate $\uparrow$ & Avg. Steps $\downarrow$ & Success Rate $\uparrow$ & Avg. Steps $\downarrow$ \\
\midrule
OmniTool (V2+GPT-4o) & 50.0 & 11.5 & 36.7 & 13.8 \\
Agent S2 (Claude-3.7-Sonnet) & 60.0 & 12.9 & 33.3 & 16.2 \\
UI-TARS-1.5 & 43.3 & 14.5 & 20.0 & 16.7 \\

\midrule
\textbf{\sysname (Qwen2.5-VL-32B)} & \textbf{83.3} & \textbf{7.1} & \textbf{76.7} & \textbf{7.5} \\
\textbf{\sysname (Qwen2.5-VL-7B)} & \underline{80.0} & \underline{7.5} & \underline{66.7} & \underline{8.6} \\
\bottomrule
\end{tabular}
\caption{\textbf{Performance comparison on OpenDCIM and EcoStruxure IT}, with both success rate and average steps reported.}
\label{tab:main}
\end{table*}

\subsection{RQ1: Task Performance across Platforms}

To answer RQ1, we evaluate our \sysname on both OpenDCIM and EcoStruxure IT, comparing its performance with three SOTA baselines: OmniTool, Agent S2, and UI-TARS-1.5. As shown in Table~\ref{tab:main}, our framework achieves the highest success rates and lowest average step counts across platforms, with particularly strong results from the Qwen2.5-VL-32B backbone and its lightweight variant Qwen2.5-VL-7B.

On OpenDCIM, our full system reaches a success rate of 83.3\% with only 7.1 average steps, and maintains comparable performance with the lightweight version. On the more complex EcoStruxure IT platform, which poses greater challenges due to less consistent interface design and deeper GUI hierarchies, our method still achieves success rates above 66.7\%, far exceeding all baselines. These results indicate its strong generalizability and adaptability across heterogeneous platforms.

\begin{figure*}[ht!]
    \centering
    \hspace{-1cm}
    \includegraphics[width=0.95\linewidth]{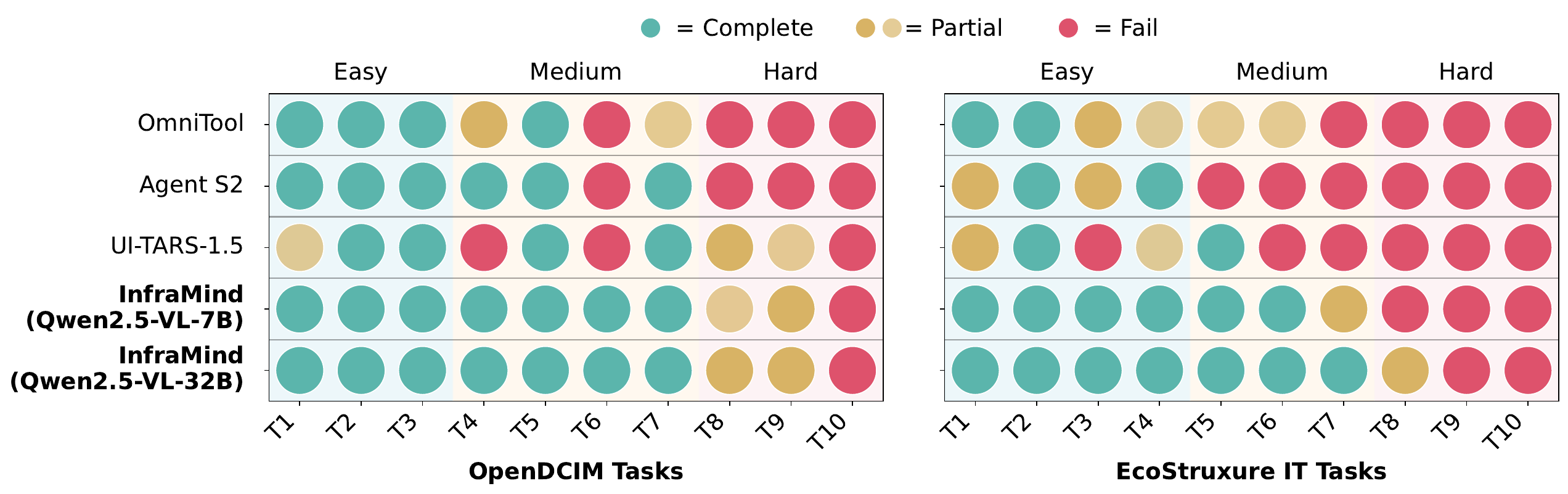}
    \caption{
\textbf{Task Completion Heatmaps}, for OpenDCIM (top) and EcoStruxure IT (bottom), across five automation frameworks. Colours denote task completion levels based on repeated trials: teal = Complete (3 out of 3 runs succeeded), ochre = Partial (1 or 2 runs succeeded), crimson = Fail (0 out of 3 runs succeeded). Tasks are grouped by difficulty (Easy, Medium, Hard).
}

    \label{fig:f4}
\end{figure*}

To further analyze task-specific performance, we visualize per-task outcomes in Figure~\ref{fig:f4}. Tasks are grouped by difficulty, and each cell represents the aggregated completion status (Complete /Partial/Fail) over three independent runs.
Across both platforms, our agent achieves high completion rates and low variance on easy and medium tasks, demonstrating consistent performance and minimal failure. This indicates effective learning of interface patterns and reliable generalization of common operations.
On hard tasks, particularly those on EcoStruxure IT, baseline methods exhibit significant degradation, often failing to complete the tasks in all three runs. In contrast, our framework shows substantially improved outcomes, with several hard tasks successfully completed or partially completed. This suggests that our exploration-centric approach, including the use of action-flow trees, state localization and reflection-based dynamic replanning, equips the system to better handle long-horizon goals and recover from intermediate uncertainties.

These findings demonstrate that our \sysname is not only strong in stable and straightforward settings, but also scales gracefully with increasing task complexity. By leveraging prior experience, dynamic reasoning, and modular collaboration, our framework achieves both a higher success rate and better execution efficiency, even under the constraints and unpredictability of real-world industrial software environments.
\\

\subsection{RQ2: Impact of Components and Model Variants}
To answer RQ2, we perform ablation experiments on OpenDCIM to assess the role of planning module, exploration module, and VLM backbones. As shown in Table \ref{tab:abs}, when the memory-driven planning component is omitted, we observe a substantial decline in success rate and efficiency, highlighting the critical importance of learning from historical experience, task planning, and dynamic adjustment. Further excluding both planning and exploration modules results in an even more pronounced drop in performance, which underscores the necessity of systematic exploration for developing a comprehensive understanding of the target software.

\begin{figure*}[ht!]
    \centering
    \hspace{0.8cm} 
    \includegraphics[width=1.0\linewidth]{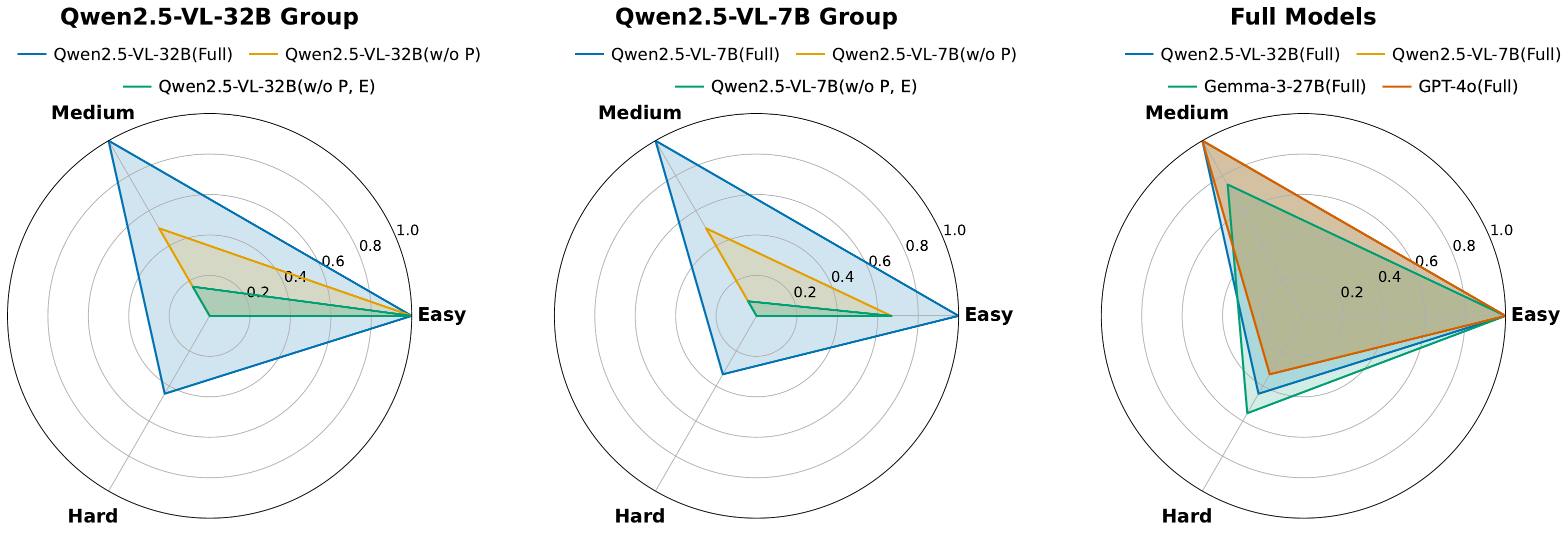}
    \caption{\textbf{Radar Plots of Task Success Rates across Difficulty Levels.} Left: Qwen2.5-VL-32B variants. Middle: Qwen2.5-VL-7B variants. Right: Full framework instantiated with different VLM backbones. \textit{P} and \textit{E} denote the \textit{Planning} and \textit{Exploration} modules respectively. The shaded area represents the normalized average success rate across tasks.}

    \label{fig:f5}
\end{figure*}

To further dissect component contributions across different task complexities, we visualise success rates by difficulty level in Figure~\ref{fig:f5}. Each radar chart shows the completion ratio for Easy, Medium, and Hard tasks across various model configurations. From the Qwen2.5-VL-32B and Qwen2.5-VL-7B groups, we observe a clear monotonic improvement as more components are included: from the base variant (w/o Planning and Exploration), to adding Exploration only, to the full system. This trend holds across all difficulty levels, but is particularly pronounced on Hard tasks. This confirms that our proposed modules jointly contribute to our \sysname ability to solve complex multi-step goals with minimal trial-and-error.

\begin{table*}[h]
\centering
\begin{tabular}{llcc}
\toprule
\multirow{2}{*}{\textbf{Model}} & \multirow{2}{*}{\textbf{Module}} & \multicolumn{2}{c}{\textbf{OpenDCIM}} \\ 
\cmidrule(lr){3-4}
& & Success Rate $\uparrow$ & Avg. Steps $\downarrow$ \\
\midrule
\multirow{3}{*}{\textbf{Qwen2.5-VL-32B}} 
    & \textbf{\sysname (Full) }    & \textbf{83.3} & \underline{7.1} \\
    & w/o Planning       & 50.0         & 11.7         \\
    & w/o Planning, Exploration       & 36.7          & 13.7         \\
\cmidrule(lr){1-4}
\multirow{3}{*}{\textbf{Qwen2.5-VL-7B}} 
    & \textbf{\sysname (Full) }      & \underline{80.0} & 7.5 \\
    & w/o Planning        & 40.0          & 13.2         \\
    & w/o Planning, Exploration        & 23.3          & 15.7         \\
\cmidrule(lr){1-4}
\multirow{1}{*}{Gemma-3-27B} 
    & \sysname (Full)                    & 76.7          & 9.6         \\
\multirow{1}{*}{GPT-4o} 
    & \sysname (Full)                     & \underline{80.0}          & \textbf{6.8}         \\
\bottomrule
\end{tabular}
\caption{\textbf{Ablation study of different models and modules on OpenDCIM.}}
\label{tab:abs}
\end{table*}

Moreover, substituting the core VLM with different backbones (Qwen2.5-VL-32B, Qwen2.5-VL-7B, Gemma-3-27B, GPT-4o) demonstrates that our framework is broadly applicable and consistently achieves strong results. Notably,  As seen in the rightmost radar chart, the fully equipped Qwen2.5-VL-7B configuration achieves comparable task performance to the much larger GPT-4o model. This highlights the efficacy of alleviating the reasoning burden of lightweight models by equipping them with planning and exploration knowledge acquired during exploration. The ability to bridge this performance gap through architecture rather than scale is critical for real-world deployment where resource constraints often preclude large-model usage.

In summary, the comparison results reinforce that our agentic design not only improves absolute performance but also expands model robustness across task complexities, and elevates lightweight models to a new operational level that brings them closer to frontier models without increasing deployment cost.


\subsection{RQ3: Case Study on Exploration-driven Efficiency and Security Mechanism}
\begin{figure*}[ht!]
    \centering
    \includegraphics[width=1.0\linewidth]{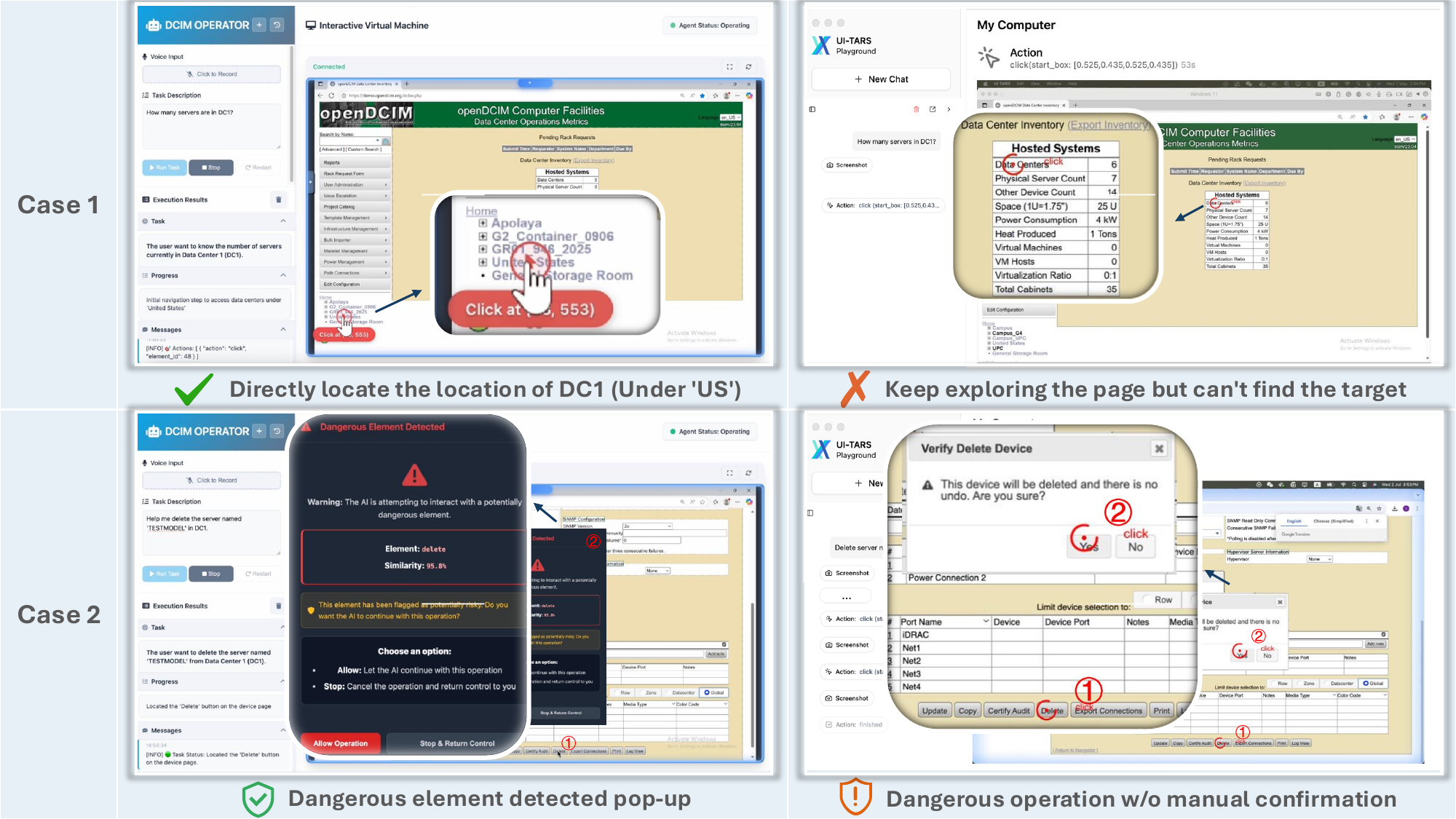}
    \caption{\textbf{Demonstration of case study results}, comparing our \sysname (left) with \textit{UI-TARS} (right) on two representative DCIM tasks. Case 1: Our agent benefits from learning-based exploration, enabling it to directly and efficiently locate the target interface, whereas UI-TARS fails to identify the correct position. Case 2: Our framework incorporates robust safety mechanisms, successfully detecting dangerous operations and prompting user confirmation, while UI-TARS proceeds with risky actions without manual safeguards.}
    \label{fig:f6}
\end{figure*}

To answer RQ3, we present representative case studies from the OpenDCIM benchmark that illustrate how our framework’s accumulated experience enables faster, more targeted task execution and ensures robust operational safety (Figure~\ref{fig:f6}).

\textbf{Case 1: Learning-driven Navigation Efficiency.}
In a typical scenario, the agent is required to locate a specific data center (“DC1” or “DC2”), accessible via the “United States” entry on the home page. Without accumulated learning, a baseline agent such as UI-TARS must exhaustively explore the entire interface, repeatedly examining elements and navigation paths in an inefficient and error-prone process. By contrast, our \sysname leverages interface-specific experience acquired through systematic exploration. As a result, it can directly identify the optimal navigation path, rapidly narrowing the search space and reaching the target interface with minimal trial and error. This case demonstrates that our learning-based approach significantly improves operational efficiency and effectiveness in complex GUI environments.


\textbf{Case 2: Built-in Safety Mechanisms.}
We further compare safety-related behaviours. When tasked with deleting a server, our \sysname incorporates robust safety modules that proactively detect dangerous UI elements and trigger a manual confirmation pop-up before executing the potentially irreversible action. This ensures that high-risk operations require explicit user validation, effectively preventing accidental or unsafe behaviour. In contrast, UI-TARS proceeds with the deletion operation without additional safeguards or manual intervention, thereby exposing the system to elevated operational risks. This case highlights the importance of integrated safety mechanisms for mission-critical applications.

\section{Related Work}

\subsection{Robotic Process Automation (RPA)}
RPA has emerged as a widely adopted solution for automating routine, rules-based tasks within enterprise information systems, aiming to enhance operational efficiency and alleviate human workload \cite{syed2020roboticRPA6,moreira2023processRPA7}. RPA technologies enable software to mimic user interactions with GUIs, and have been embraced across various sectors including finance, insurance, and science industry\cite{lamberton2017impactRPA8,9001110RPA9}. Major commercial RPA platforms, such as UiPath, Blue Prism, and Automation Anywhere, have accelerated industry uptake, offering comprehensive toolsets for task automation, monitoring, and management \cite{9001110RPA9}.

Specifically, RPA has been applied effectively across various domains to enhance operational efficiency and automate complex tasks. ~\citet{en13092342RPA1} developed an energy-saving operations system for remote operation of building management equipment, leveraging RPA to optimize energy use, achieving significant energy and CO2 reductions in real-world trials within a large shopping mall. In the power grid dispatching context, ~\citet{LiuRPA2} implemented an RPA-based system utilizing UiBot and WPS tables to automate data reporting tasks, significantly reducing manual workload and streamlining data handling processes. ~\citet{Chen2025RPA3} proposed an RPA-driven framework combined with Grey Wolf Optimization for smart microgrid management, achieving reductions in emissions, operational costs, and increased power supply reliability. ~\citet{en14165191RPA4} explored the application of RPA in smart city initiatives, specifically to automate the management of electricity billing documents at Bydgoszcz City Hall, demonstrating measurable cost efficiency and operational improvements. ~\citet{app15020854RPA5} introduced an integrated sustainable model that combined RPA and machine learning to enhance predictive maintenance, substantially improving maintenance metrics such as mean time between failures and unplanned downtime. 


Recent studies highlight significant benefits of Robotic Process Automation (RPA), including cost reduction, accelerated processes, lower error rates, and improved data quality \cite{moreira2023processRPA7,lamberton2017impactRPA8, lacity2016newRPA10}. However, substantial challenges remain. In particular, the suitability of workflows for automation remains limited by their inherent complexity and variability \cite{moreira2023processRPA7, 9001110RPA9}. Furthermore, implementing and scaling RPA requires considerable manual effort, as bots require detailed configuration and frequent updates to manage minor interface changes \cite{lamberton2017impactRPA8,leno2018multiRPA11}.

In summary, while RPA has delivered significant practical value in automating repetitive business processes, its dependence on manual scripting, limited adaptability to dynamic software environments, and lack of intelligent process understanding pose substantial barriers in complex mission-critical domains such as industrial management software. These shortcomings motivate the exploration of more adaptive and learning-based automation paradigms, as addressed in our proposed framework.


\subsection{GUI Agents}
Recent advances in large language models (LLMs) have spurred rapid progress in autonomous GUI agents, with research targeting general-purpose automation in diverse desktop, web, and mobile environments \cite{zhao2025worldguiinteractivebenchmarkdesktop,wu-etal-2025-webwalker,hu-etal-2025-osagents}. These agents, typically designed for open-ended digital tasks, demonstrate the ability to perceive coplex graphical interfaces, ground actions to UI elements, plan over long horizons, and interact through human-like sequences of clicks and keystrokes. State-of-the-art frameworks such as UI-TARS \cite{qin2025uitars}, Agent S/S2 \cite{Agent-S, Agent-S2}, and WorldGUI-Agent \cite{zhao2025worldguiinteractivebenchmarkdesktop} exemplify this trend, achieving strong performance across a range of benchmark suites, including OSWorld \cite{xie2024osworld}, WindowsAgentArena \cite{bonatti2024windowsagentarena} and ScreenSpot-Pro \cite{li2025screenspotpro}.

Recent works have enhanced planning robustness with mechanisms such as stepwise plan verification \cite{zhao2025worldguiinteractivebenchmarkdesktop}, error detection with backtracking \cite{wu2025backtrackagent}, and explicit rollback during web navigation \cite{zhang2025enhancing}, aiming to address the brittleness of open-ended agents in dynamic or unfamiliar environments. Autonomous exploration has also emerged as a key research theme: frameworks like WebWalker \cite{wu-etal-2025-webwalker} and Explorer \cite{pahuja2025explorer} employ multi-agent exploration or trajectory synthesis to systematically traverse web or GUI spaces, improving task coverage and data collection for further model training.

Despite these advances, most existing works maintain a general-purpose focus, seeking universal agents capable of operating “in the wild” but often lacking domain-specific adaptation. GUI grounding models, such as those surveyed in \cite{fan2025guibeealignguiaction, singh2025trishul}, struggle with cross-environment generalization, requiring frequent re-training or manual intervention in novel or highly-specialized software contexts. While some frameworks introduce few-shot learning from demonstrations \cite{liu2025learnact} or environment-free exploration \cite{zheng2025vemenvironmentfreeexplorationtraining}, robust domain mastery and safe operation in mission-critical industrial software remain underexplored.

To sum up, while existing LLM-based GUI agents achieve impressive results in open-domain tasks, they lack domain-specific adaptation and robust recovery for mission-critical industrial software. Our work addresses these gaps with a learning-centric framework tailored for complex industrial environments.


\section{Discussion}
Despite demonstrating strong performance across diverse industrial scenarios, our framework has several limitations that present opportunities for future improvement. We categorise these along three major dimensions: perception, exploration, and deployment scope.

\noindent\textbf{Stronger Perception Capability} The agent's performance is fundamentally dependent on its perception capabilities. While the current OmniParser based method, leveraging pre-trained object detection models such as YOLO, has proven effective for a range of industry scenarios, it suffers from inherent stability issues and incomplete coverage of GUI elements. This reliance on a predefined model can lead to instances where buttons, text boxes, or other critical components are not accurately identified. Recent work has proposed specialised methods to improve GUI perception accuracy, including more robust multimodal alignment and adaptive component recognition \cite{tang2025guigrounding,singh2025trishul}. A more robust solution lies in developing a direct, coordinate based localization model. However, this approach is currently limited by the spatial imprecision and contextual shortcomings of SOTA VLMs when interpreting complex interfaces. Future work will focus on improving VLM based localization through advanced fine-tuning techniques and novel model architectures designed for precise, pixel level GUI element identification. This will enable the agent to interact with a wider variety of interfaces with greater accuracy and reliability.




\noindent\textbf{More Efficient Exploration} Another key challenge lies in the task-specific exploration phase, particularly in long-horizon or low-reward tasks. Currently, the agent may spend significant time rediscovering viable execution paths, especially when initial trials fail due to limited semantic understanding or ambiguous GUI layouts.
Recent studies have attempted to enhance GUI agents with reinforcement learning (RL), enabling them to learn exploration policies over time \cite{lu2025uiR1,zheng2025vemenvironmentfreeexplorationtraining, luo2025guir1}. However, applying RL to GUI automation poses a fundamental challenge: most GUI environments are non-reversible, which means incorrect actions can permanently change the system state, making recovery or policy refinement difficult.
To address this, we propose incorporating virtual machine-based rollback mechanisms, enabling the agent to safely explore and retry within a sandboxed environment. By capturing and restoring exact GUI states, this approach mitigates the irreversibility problem and opens the door to applying RL techniques for long-horizon planning and continual learning. In future work, we plan to integrate VM-based rollback with policy learning strategies to further optimise success probability while reducing redundant trials.

\noindent\textbf{Wider Application and Actual Deployment
} Currently, our framework is validated on two representative DCIM systems, which reflect core operational tasks in data center management. However, these platforms cover only a subset of the software heterogeneity observed in industry. Future research will expand the benchmark suite to include broader classes of management software (e.g., power systems, HVAC interfaces, network control panels), thus challenging the framework to generalise across domains with different GUI styles, semantics, and user workflows.
In addition, we will prioritise real-world deployment and operator feedback, enabling us to iteratively refine the agent’s capabilities under production constraints. These deployments will help uncover long-tail behaviours, reliability issues, and safety concerns that are not fully captured in simulation. Integrating such feedback into our training and evaluation pipeline will be essential for building practically useful, high-stakes automation systems.


\section{Conclusion}

Mission-critical infrastructure relies on complex industrial management software, yet both traditional RPA and generic LLM-based agents struggle to meet the sector’s dynamic and safety-critical requirements. To address these challenges, we propose a learning-based agentic automation framework \sysname, enabling autonomous GUI exploration and operation via a modular Perception–Reasoning–Action loop and virtual machine sandboxing. Extensive experiments on benchmarks and real-world data center deployments confirm the framework’s effectiveness, scalability, and resilience to interface changes, paving the way for future research on broader industrial applications, enhanced model efficiency, and deeper domain integration.

\bibliographystyle{ACM-Reference-Format}
\bibliography{reference}


\begin{thebibliography}{55}


\ifx \showCODEN    \undefined \def \showCODEN     #1{\unskip}     \fi
\ifx \showDOI      \undefined \def \showDOI       #1{#1}\fi
\ifx \showISBNx    \undefined \def \showISBNx     #1{\unskip}     \fi
\ifx \showISBNxiii \undefined \def \showISBNxiii  #1{\unskip}     \fi
\ifx \showISSN     \undefined \def \showISSN      #1{\unskip}     \fi
\ifx \showLCCN     \undefined \def \showLCCN      #1{\unskip}     \fi
\ifx \shownote     \undefined \def \shownote      #1{#1}          \fi
\ifx \showarticletitle \undefined \def \showarticletitle #1{#1}   \fi
\ifx \showURL      \undefined \def \showURL       {\relax}        \fi
\providecommand\bibfield[2]{#2}
\providecommand\bibinfo[2]{#2}
\providecommand\natexlab[1]{#1}
\providecommand\showeprint[2][]{arXiv:#2}

\bibitem[AG(2021)]%
        {siemens1-2021claritylc}
\bibfield{author}{\bibinfo{person}{Siemens AG}.} \bibinfo{year}{2021}\natexlab{}.
\newblock \bibinfo{title}{Data Center Clarity LC: Technical Sheet}.
\newblock \bibinfo{howpublished}{\url{https://assets.new.siemens.com/siemens/assets/api/uuid:39138c75c58abd0d3def12cca7164675a335bf16/siemens-datacenter-clarity-lc-tech-sheet-2021.pdf}}.
\newblock


\bibitem[Agashe et~al\mbox{.}(2025a)]%
        {Agent-S}
\bibfield{author}{\bibinfo{person}{Saaket Agashe}, \bibinfo{person}{Jiuzhou Han}, \bibinfo{person}{Shuyu Gan}, \bibinfo{person}{Jiachen Yang}, \bibinfo{person}{Ang Li}, {and} \bibinfo{person}{Xin~Eric Wang}.} \bibinfo{year}{2025}\natexlab{a}.
\newblock \showarticletitle{{Agent S: An Open Agentic Framework that Uses Computers Like a Human}}. In \bibinfo{booktitle}{\emph{International Conference on Learning Representations (ICLR)}}.
\newblock
\urldef\tempurl%
\url{https://arxiv.org/abs/2410.08164}
\showURL{%
\tempurl}


\bibitem[Agashe et~al\mbox{.}(2025b)]%
        {Agent-S2}
\bibfield{author}{\bibinfo{person}{Saaket Agashe}, \bibinfo{person}{Kyle Wong}, \bibinfo{person}{Vincent Tu}, \bibinfo{person}{Jiachen Yang}, \bibinfo{person}{Ang Li}, {and} \bibinfo{person}{Xin~Eric Wang}.} \bibinfo{year}{2025}\natexlab{b}.
\newblock \bibinfo{title}{Agent S2: A Compositional Generalist-Specialist Framework for Computer Use Agents}.
\newblock
\newblock
\showeprint[arxiv]{2504.00906}~[cs.AI]
\urldef\tempurl%
\url{https://arxiv.org/abs/2504.00906}
\showURL{%
\tempurl}


\bibitem[Bonatti et~al\mbox{.}(2024)]%
        {bonatti2024windowsagentarena}
\bibfield{author}{\bibinfo{person}{Rogerio Bonatti}, \bibinfo{person}{Dan Zhao}, \bibinfo{person}{Francesco Bonacci}, \bibinfo{person}{Dillon Dupont}, \bibinfo{person}{Sara Abdali}, \bibinfo{person}{Yinheng Li}, \bibinfo{person}{Yadong Lu}, \bibinfo{person}{Justin Wagle}, \bibinfo{person}{Kazuhito Koishida}, \bibinfo{person}{Arthur Bucker}, {et~al\mbox{.}}} \bibinfo{year}{2024}\natexlab{}.
\newblock \showarticletitle{Windows agent arena: Evaluating multi-modal os agents at scale}.
\newblock \bibinfo{journal}{\emph{arXiv preprint arXiv:2409.08264}} (\bibinfo{year}{2024}).
\newblock


\bibitem[Brown et~al\mbox{.}(2024)]%
        {brown2024self-eval}
\bibfield{author}{\bibinfo{person}{Hannah Brown}, \bibinfo{person}{Leon Lin}, \bibinfo{person}{Kenji Kawaguchi}, {and} \bibinfo{person}{Michael Shieh}.} \bibinfo{year}{2024}\natexlab{}.
\newblock \showarticletitle{Self-evaluation as a defense against adversarial attacks on llms}.
\newblock \bibinfo{journal}{\emph{arXiv preprint arXiv:2407.03234}} (\bibinfo{year}{2024}).
\newblock


\bibitem[Chawla et~al\mbox{.}(2024)]%
        {chawla2024guideRPAweak}
\bibfield{author}{\bibinfo{person}{Rajat Chawla}, \bibinfo{person}{Adarsh Jha}, \bibinfo{person}{Muskaan Kumar}, \bibinfo{person}{Mukunda NS}, {and} \bibinfo{person}{Ishaan Bhola}.} \bibinfo{year}{2024}\natexlab{}.
\newblock \showarticletitle{Guide: Graphical user interface data for execution}.
\newblock \bibinfo{journal}{\emph{arXiv preprint arXiv:2404.16048}} (\bibinfo{year}{2024}).
\newblock


\bibitem[Chen et~al\mbox{.}(2025)]%
        {Chen2025RPA3}
\bibfield{author}{\bibinfo{person}{Bin Chen}, \bibinfo{person}{Zeke Li}, \bibinfo{person}{Bijing Liu}, \bibinfo{person}{Haiwei Fan}, {and} \bibinfo{person}{Qiutian Zhong}.} \bibinfo{year}{2025}\natexlab{}.
\newblock \showarticletitle{Robust optimization for smart demand side management in microgrids using robotic process automation and grey wolf optimization}.
\newblock \bibinfo{journal}{\emph{Scientific Reports}} \bibinfo{volume}{15}, \bibinfo{number}{1} (\bibinfo{year}{2025}), \bibinfo{pages}{19440}.
\newblock
\showISSN{2045-2322}
\urldef\tempurl%
\url{https://doi.org/10.1038/s41598-025-03728-8}
\showDOI{\tempurl}


\bibitem[De~Bruijne and Van~Eeten(2007)]%
        {de2007systemsMCI1}
\bibfield{author}{\bibinfo{person}{Mark De~Bruijne} {and} \bibinfo{person}{Michel Van~Eeten}.} \bibinfo{year}{2007}\natexlab{}.
\newblock \showarticletitle{Systems that should have failed: critical infrastructure protection in an institutionally fragmented environment}.
\newblock \bibinfo{journal}{\emph{Journal of contingencies and crisis management}} \bibinfo{volume}{15}, \bibinfo{number}{1} (\bibinfo{year}{2007}), \bibinfo{pages}{18--29}.
\newblock


\bibitem[Electric(2025)]%
        {schneider3-2025_complexitiesGridMod}
\bibfield{author}{\bibinfo{person}{Schneider Electric}.} \bibinfo{year}{2025}\natexlab{}.
\newblock \bibinfo{title}{Complexities of Grid Modernization}.
\newblock \bibinfo{howpublished}{\url{https://www.se.com/us/en/download/document/Complexitie_Grid_Modernization/}}.
\newblock


\bibitem[Enríquez et~al\mbox{.}(2020)]%
        {9001110RPA9}
\bibfield{author}{\bibinfo{person}{J.~G. Enríquez}, \bibinfo{person}{A. Jiménez-Ramírez}, \bibinfo{person}{F.~J. Domínguez-Mayo}, {and} \bibinfo{person}{J.~A. García-García}.} \bibinfo{year}{2020}\natexlab{}.
\newblock \showarticletitle{Robotic Process Automation: A Scientific and Industrial Systematic Mapping Study}.
\newblock \bibinfo{journal}{\emph{IEEE Access}}  \bibinfo{volume}{8} (\bibinfo{year}{2020}), \bibinfo{pages}{39113--39129}.
\newblock
\urldef\tempurl%
\url{https://doi.org/10.1109/ACCESS.2020.2974934}
\showDOI{\tempurl}


\bibitem[Fan et~al\mbox{.}(2025)]%
        {fan2025guibeealignguiaction}
\bibfield{author}{\bibinfo{person}{Yue Fan}, \bibinfo{person}{Handong Zhao}, \bibinfo{person}{Ruiyi Zhang}, \bibinfo{person}{Yu Shen}, \bibinfo{person}{Xin~Eric Wang}, {and} \bibinfo{person}{Gang Wu}.} \bibinfo{year}{2025}\natexlab{}.
\newblock \bibinfo{title}{GUI-Bee: Align GUI Action Grounding to Novel Environments via Autonomous Exploration}.
\newblock
\newblock
\showeprint[arxiv]{2501.13896}~[cs.CL]
\urldef\tempurl%
\url{https://arxiv.org/abs/2501.13896}
\showURL{%
\tempurl}


\bibitem[Heslin(2019)]%
        {uptime8-2019_tryharder}
\bibfield{author}{\bibinfo{person}{Kevin Heslin}.} \bibinfo{year}{2019}\natexlab{}.
\newblock \bibinfo{title}{How to avoid outages: Try harder!}
\newblock \bibinfo{howpublished}{\url{https://journal.uptimeinstitute.com/how-to-avoid-outages-try-harder/}}.
\newblock


\bibitem[Houliotis et~al\mbox{.}(2018)]%
        {houliotis2018missionMCI7}
\bibfield{author}{\bibinfo{person}{Kyriakos Houliotis}, \bibinfo{person}{Panagiotis Oikonomidis}, \bibinfo{person}{Periklis Charchalakis}, {and} \bibinfo{person}{Elias Stipidis}.} \bibinfo{year}{2018}\natexlab{}.
\newblock \showarticletitle{Mission-critical systems design framework}.
\newblock \bibinfo{journal}{\emph{Advances in Science, Technology and Engineering Systems Journal}} \bibinfo{volume}{3}, \bibinfo{number}{2} (\bibinfo{year}{2018}), \bibinfo{pages}{128--137}.
\newblock


\bibitem[Hu et~al\mbox{.}(2025)]%
        {hu-etal-2025-osagents}
\bibfield{author}{\bibinfo{person}{Xueyu Hu}, \bibinfo{person}{Tao Xiong}, \bibinfo{person}{Biao Yi}, \bibinfo{person}{Zishu Wei}, \bibinfo{person}{Ruixuan Xiao}, \bibinfo{person}{Yurun Chen}, \bibinfo{person}{Jiasheng Ye}, \bibinfo{person}{Meiling Tao}, \bibinfo{person}{Xiangxin Zhou}, \bibinfo{person}{Ziyu Zhao}, \bibinfo{person}{Yuhuai Li}, \bibinfo{person}{Shengze Xu}, \bibinfo{person}{Shenzhi Wang}, \bibinfo{person}{Xinchen Xu}, \bibinfo{person}{Shuofei Qiao}, \bibinfo{person}{Zhaokai Wang}, \bibinfo{person}{Kun Kuang}, \bibinfo{person}{Tieyong Zeng}, \bibinfo{person}{Liang Wang}, \bibinfo{person}{Jiwei Li}, \bibinfo{person}{Yuchen~Eleanor Jiang}, \bibinfo{person}{Wangchunshu Zhou}, \bibinfo{person}{Guoyin Wang}, \bibinfo{person}{Keting Yin}, \bibinfo{person}{Zhou Zhao}, \bibinfo{person}{Hongxia Yang}, \bibinfo{person}{Fan Wu}, \bibinfo{person}{Shengyu Zhang}, {and} \bibinfo{person}{Fei Wu}.} \bibinfo{year}{2025}\natexlab{}.
\newblock \showarticletitle{{OS} Agents: A Survey on {MLLM}-based Agents for Computer, Phone and Browser Use}. In \bibinfo{booktitle}{\emph{Proceedings of the 63rd Annual Meeting of the Association for Computational Linguistics (Volume 1: Long Papers)}}, \bibfield{editor}{\bibinfo{person}{Wanxiang Che}, \bibinfo{person}{Joyce Nabende}, \bibinfo{person}{Ekaterina Shutova}, {and} \bibinfo{person}{Mohammad~Taher Pilehvar}} (Eds.). \bibinfo{publisher}{Association for Computational Linguistics}, \bibinfo{address}{Vienna, Austria}, \bibinfo{pages}{7436--7465}.
\newblock
\showISBNx{979-8-89176-251-0}
\urldef\tempurl%
\url{https://aclanthology.org/2025.acl-long.369/}
\showURL{%
\tempurl}


\bibitem[Huang(2021)]%
        {huang2021dataDCIM}
\bibfield{author}{\bibinfo{person}{Dongmei Huang}.} \bibinfo{year}{2021}\natexlab{}.
\newblock \showarticletitle{Data center infrastructure management}.
\newblock \bibinfo{journal}{\emph{Data Center Handbook: Plan, Design, Build, and Operations of a Smart Data Center}} (\bibinfo{year}{2021}), \bibinfo{pages}{627--644}.
\newblock


\bibitem[Institute(2023)]%
        {uptime7-2023StaffingSurvey}
\bibfield{author}{\bibinfo{person}{Uptime Institute}.} \bibinfo{year}{2023}\natexlab{}.
\newblock \bibinfo{title}{Operators Struggle to Overcome Ongoing Staff and Skills Shortage}.
\newblock \bibinfo{howpublished}{\url{https://datacenter.uptimeinstitute.com/rs/711-RIA-145/images/2023StaffingSurvey.Report.12152023.pdf}}.
\newblock


\bibitem[Jang et~al\mbox{.}(2025)]%
        {jang2025vl2lite}
\bibfield{author}{\bibinfo{person}{Jinseong Jang}, \bibinfo{person}{Chunfei Ma}, {and} \bibinfo{person}{Byeongwon Lee}.} \bibinfo{year}{2025}\natexlab{}.
\newblock \showarticletitle{VL2Lite: Task-Specific Knowledge Distillation from Large Vision-Language Models to Lightweight Networks}. In \bibinfo{booktitle}{\emph{Proceedings of the Computer Vision and Pattern Recognition Conference}}. \bibinfo{pages}{30073--30083}.
\newblock


\bibitem[Johansson et~al\mbox{.}(2013)]%
        {JOHANSSON201327MCI2}
\bibfield{author}{\bibinfo{person}{Jonas Johansson}, \bibinfo{person}{Henrik Hassel}, {and} \bibinfo{person}{Enrico Zio}.} \bibinfo{year}{2013}\natexlab{}.
\newblock \showarticletitle{Reliability and vulnerability analyses of critical infrastructures: Comparing two approaches in the context of power systems}.
\newblock \bibinfo{journal}{\emph{Reliability Engineering \& System Safety}}  \bibinfo{volume}{120} (\bibinfo{year}{2013}), \bibinfo{pages}{27--38}.
\newblock
\showISSN{0951-8320}
\urldef\tempurl%
\url{https://doi.org/10.1016/j.ress.2013.02.027}
\showDOI{\tempurl}


\bibitem[Lacity and Willcocks(2016)]%
        {lacity2016newRPA10}
\bibfield{author}{\bibinfo{person}{Mary~C Lacity} {and} \bibinfo{person}{Leslie~P Willcocks}.} \bibinfo{year}{2016}\natexlab{}.
\newblock \showarticletitle{A new approach to automating services}.
\newblock \bibinfo{journal}{\emph{MIT Sloan Management Review}} (\bibinfo{year}{2016}).
\newblock


\bibitem[Lamberton et~al\mbox{.}(2017)]%
        {lamberton2017impactRPA8}
\bibfield{author}{\bibinfo{person}{Chris Lamberton}, \bibinfo{person}{Damiano Brigo}, {and} \bibinfo{person}{Dave Hoy}.} \bibinfo{year}{2017}\natexlab{}.
\newblock \showarticletitle{Impact of Robotics, RPA and AI on the insurance industry: challenges and opportunities}.
\newblock \bibinfo{journal}{\emph{Journal of Financial Perspectives}} \bibinfo{volume}{4}, \bibinfo{number}{1} (\bibinfo{year}{2017}).
\newblock


\bibitem[Leno et~al\mbox{.}(2018)]%
        {leno2018multiRPA11}
\bibfield{author}{\bibinfo{person}{Volodymyr Leno}, \bibinfo{person}{Marlon Dumas}, \bibinfo{person}{Fabrizio~Maria Maggi}, {and} \bibinfo{person}{Marcello La~Rosa}.} \bibinfo{year}{2018}\natexlab{}.
\newblock \showarticletitle{Multi-perspective process model discovery for robotic process automation}. In \bibinfo{booktitle}{\emph{Proceedings of the Doctoral Consortium Papers Presented at the 30th International Conference on Advanced Information Systems Engineering (CAiSE 2018)}}, Vol.~\bibinfo{volume}{2114}. CEUR-WS, \bibinfo{pages}{37--45}.
\newblock


\bibitem[Li et~al\mbox{.}(2025)]%
        {li2025screenspotpro}
\bibfield{author}{\bibinfo{person}{Kaixin Li}, \bibinfo{person}{Meng Ziyang}, \bibinfo{person}{Hongzhan Lin}, \bibinfo{person}{Ziyang Luo}, \bibinfo{person}{Yuchen Tian}, \bibinfo{person}{Jing Ma}, \bibinfo{person}{Zhiyong Huang}, {and} \bibinfo{person}{Tat-Seng Chua}.} \bibinfo{year}{2025}\natexlab{}.
\newblock \showarticletitle{ScreenSpot-Pro: {GUI} Grounding for Professional High-Resolution Computer Use}. In \bibinfo{booktitle}{\emph{Workshop on Reasoning and Planning for Large Language Models}}.
\newblock
\urldef\tempurl%
\url{https://openreview.net/forum?id=XaKNDIAHas}
\showURL{%
\tempurl}


\bibitem[Liu et~al\mbox{.}(2025)]%
        {liu2025learnact}
\bibfield{author}{\bibinfo{person}{Guangyi Liu}, \bibinfo{person}{Pengxiang Zhao}, \bibinfo{person}{Liang Liu}, \bibinfo{person}{Zhiming Chen}, \bibinfo{person}{Yuxiang Chai}, \bibinfo{person}{Shuai Ren}, \bibinfo{person}{Hao Wang}, \bibinfo{person}{Shibo He}, {and} \bibinfo{person}{Wenchao Meng}.} \bibinfo{year}{2025}\natexlab{}.
\newblock \showarticletitle{Learnact: Few-shot mobile gui agent with a unified demonstration benchmark}.
\newblock \bibinfo{journal}{\emph{arXiv preprint arXiv:2504.13805}} (\bibinfo{year}{2025}).
\newblock


\bibitem[Liu et~al\mbox{.}(2023)]%
        {LiuRPA2}
\bibfield{author}{\bibinfo{person}{Shi Liu}, \bibinfo{person}{Xuesong Qi}, {and} \bibinfo{person}{Huanqi Li}.} \bibinfo{year}{2023}\natexlab{}.
\newblock \showarticletitle{Practice of Robot Process Automation in Power Grid Dispatching Report}. In \bibinfo{booktitle}{\emph{Proceedings of the 2022 4th International Conference on Robotics, Intelligent Control and Artificial Intelligence}} (Dongguan, China) \emph{(\bibinfo{series}{RICAI '22})}. \bibinfo{publisher}{Association for Computing Machinery}, \bibinfo{address}{New York, NY, USA}, \bibinfo{pages}{212–216}.
\newblock
\showISBNx{9781450398343}
\urldef\tempurl%
\url{https://doi.org/10.1145/3584376.3584415}
\showDOI{\tempurl}


\bibitem[Lu et~al\mbox{.}(2024)]%
        {lu2024omniparserpurevisionbased}
\bibfield{author}{\bibinfo{person}{Yadong Lu}, \bibinfo{person}{Jianwei Yang}, \bibinfo{person}{Yelong Shen}, {and} \bibinfo{person}{Ahmed Awadallah}.} \bibinfo{year}{2024}\natexlab{}.
\newblock \bibinfo{title}{OmniParser for Pure Vision Based GUI Agent}.
\newblock
\newblock
\showeprint[arxiv]{2408.00203}~[cs.CV]
\urldef\tempurl%
\url{https://arxiv.org/abs/2408.00203}
\showURL{%
\tempurl}


\bibitem[Lu et~al\mbox{.}(2025)]%
        {lu2025uiR1}
\bibfield{author}{\bibinfo{person}{Zhengxi Lu}, \bibinfo{person}{Yuxiang Chai}, \bibinfo{person}{Yaxuan Guo}, \bibinfo{person}{Xi Yin}, \bibinfo{person}{Liang Liu}, \bibinfo{person}{Hao Wang}, \bibinfo{person}{Han Xiao}, \bibinfo{person}{Shuai Ren}, \bibinfo{person}{Guanjing Xiong}, {and} \bibinfo{person}{Hongsheng Li}.} \bibinfo{year}{2025}\natexlab{}.
\newblock \showarticletitle{UI-R1: Enhancing Efficient Action Prediction of GUI Agents by Reinforcement Learning}.
\newblock \bibinfo{journal}{\emph{arXiv preprint arXiv:2503.21620}} (\bibinfo{year}{2025}).
\newblock


\bibitem[Luo et~al\mbox{.}(2025)]%
        {luo2025guir1}
\bibfield{author}{\bibinfo{person}{Run Luo}, \bibinfo{person}{Lu Wang}, \bibinfo{person}{Wanwei He}, {and} \bibinfo{person}{Xiaobo Xia}.} \bibinfo{year}{2025}\natexlab{}.
\newblock \showarticletitle{Gui-r1: A generalist r1-style vision-language action model for gui agents}.
\newblock \bibinfo{journal}{\emph{arXiv preprint arXiv:2504.10458}} (\bibinfo{year}{2025}).
\newblock


\bibitem[McDaniels et~al\mbox{.}(2007)]%
        {mcdaniels2007empiricalMCI3}
\bibfield{author}{\bibinfo{person}{Timothy McDaniels}, \bibinfo{person}{Stephanie Chang}, \bibinfo{person}{Krista Peterson}, \bibinfo{person}{Joey Mikawoz}, {and} \bibinfo{person}{Dorothy Reed}.} \bibinfo{year}{2007}\natexlab{}.
\newblock \showarticletitle{Empirical framework for characterizing infrastructure failure interdependencies}.
\newblock \bibinfo{journal}{\emph{Journal of Infrastructure Systems}} \bibinfo{volume}{13}, \bibinfo{number}{3} (\bibinfo{year}{2007}), \bibinfo{pages}{175--184}.
\newblock


\bibitem[Moreira et~al\mbox{.}(2023)]%
        {moreira2023processRPA7}
\bibfield{author}{\bibinfo{person}{S{\'\i}lvia Moreira}, \bibinfo{person}{Henrique~S Mamede}, {and} \bibinfo{person}{Arnaldo Santos}.} \bibinfo{year}{2023}\natexlab{}.
\newblock \showarticletitle{Process automation using RPA--a literature review}.
\newblock \bibinfo{journal}{\emph{Procedia Computer Science}}  \bibinfo{volume}{219} (\bibinfo{year}{2023}), \bibinfo{pages}{244--254}.
\newblock


\bibitem[Müller and Žunič(2024)]%
        {browser_use2024}
\bibfield{author}{\bibinfo{person}{Magnus Müller} {and} \bibinfo{person}{Gregor Žunič}.} \bibinfo{year}{2024}\natexlab{}.
\newblock \bibinfo{booktitle}{\emph{Browser Use: Enable AI to control your browser}}.
\newblock
\urldef\tempurl%
\url{https://github.com/browser-use/browser-use}
\showURL{%
\tempurl}


\bibitem[Nguyen et~al\mbox{.}(2019)]%
        {8783313crosssystem}
\bibfield{author}{\bibinfo{person}{Van~Hoa Nguyen}, \bibinfo{person}{Tung Lam~Nguyen}, \bibinfo{person}{Quoc~Tuan Tran}, \bibinfo{person}{Yvon Besanger}, {and} \bibinfo{person}{Raphael Caire}.} \bibinfo{year}{2019}\natexlab{}.
\newblock \showarticletitle{Integration of SCADA services in cross-infrastructure holistic tests of cyber-physical energy systems}. In \bibinfo{booktitle}{\emph{2019 IEEE International Conference on Environment and Electrical Engineering and 2019 IEEE Industrial and Commercial Power Systems Europe (EEEIC / I\&CPS Europe)}}. \bibinfo{pages}{1--5}.
\newblock
\urldef\tempurl%
\url{https://doi.org/10.1109/EEEIC.2019.8783313}
\showDOI{\tempurl}


\bibitem[Ojo et~al\mbox{.}(2025)]%
        {en18133576newMCI2}
\bibfield{author}{\bibinfo{person}{Kayode~Ebenezer Ojo}, \bibinfo{person}{Akshay~Kumar Saha}, {and} \bibinfo{person}{Viranjay~Mohan Srivastava}.} \bibinfo{year}{2025}\natexlab{}.
\newblock \showarticletitle{Microgrids’ Control Strategies and Real-Time Monitoring Systems: A Comprehensive Review}.
\newblock \bibinfo{journal}{\emph{Energies}} \bibinfo{volume}{18}, \bibinfo{number}{13} (\bibinfo{year}{2025}).
\newblock
\showISSN{1996-1073}
\urldef\tempurl%
\url{https://doi.org/10.3390/en18133576}
\showDOI{\tempurl}


\bibitem[Onwujekwe and Weistroffer(2025)]%
        {1c89305696a346e386c44c89c8fff229dds}
\bibfield{author}{\bibinfo{person}{Gerald Onwujekwe} {and} \bibinfo{person}{\{Heinz Roland\} Weistroffer}.} \bibinfo{year}{2025}\natexlab{}.
\newblock \showarticletitle{Intelligent Decision Support Systems: An Analysis of the Literature and a Framework for Development}.
\newblock \bibinfo{journal}{\emph{Information Systems Frontiers}} (\bibinfo{year}{2025}).
\newblock
\showISSN{1387-3326}
\urldef\tempurl%
\url{https://doi.org/10.1007/s10796-024-10571-1}
\showDOI{\tempurl}
\newblock
\shownote{Publisher Copyright: {\textcopyright} The Author(s), under exclusive licence to Springer Science+Business Media, LLC, part of Springer Nature 2025.}.


\bibitem[{OpenDCIM Community}(2025)]%
        {opendcim}
\bibfield{author}{\bibinfo{person}{{OpenDCIM Community}}.} \bibinfo{year}{2025}\natexlab{}.
\newblock \bibinfo{booktitle}{\emph{OpenDCIM: Open Source Data Center Infrastructure Management}}.
\newblock
\urldef\tempurl%
\url{https://www.opendcim.org/}
\showURL{%
\tempurl}


\bibitem[Pahuja et~al\mbox{.}(2025)]%
        {pahuja2025explorer}
\bibfield{author}{\bibinfo{person}{Vardaan Pahuja}, \bibinfo{person}{Yadong Lu}, \bibinfo{person}{Corby Rosset}, \bibinfo{person}{Boyu Gou}, \bibinfo{person}{Arindam Mitra}, \bibinfo{person}{Spencer Whitehead}, \bibinfo{person}{Yu Su}, {and} \bibinfo{person}{Ahmed Awadallah}.} \bibinfo{year}{2025}\natexlab{}.
\newblock \showarticletitle{Explorer: Scaling exploration-driven web trajectory synthesis for multimodal web agents}.
\newblock \bibinfo{journal}{\emph{arXiv preprint arXiv:2502.11357}} (\bibinfo{year}{2025}).
\newblock


\bibitem[Patrício et~al\mbox{.}(2025)]%
        {app15020854RPA5}
\bibfield{author}{\bibinfo{person}{Leonel Patrício}, \bibinfo{person}{Leonilde Varela}, {and} \bibinfo{person}{Zilda Silveira}.} \bibinfo{year}{2025}\natexlab{}.
\newblock \showarticletitle{Proposal for a Sustainable Model for Integrating Robotic Process Automation and Machine Learning in Failure Prediction and Operational Efficiency in Predictive Maintenance}.
\newblock \bibinfo{journal}{\emph{Applied Sciences}} \bibinfo{volume}{15}, \bibinfo{number}{2} (\bibinfo{year}{2025}).
\newblock
\showISSN{2076-3417}
\urldef\tempurl%
\url{https://doi.org/10.3390/app15020854}
\showDOI{\tempurl}


\bibitem[Qin et~al\mbox{.}(2025)]%
        {qin2025uitars}
\bibfield{author}{\bibinfo{person}{Yujia Qin}, \bibinfo{person}{Yining Ye}, \bibinfo{person}{Junjie Fang}, \bibinfo{person}{Haoming Wang}, \bibinfo{person}{Shihao Liang}, \bibinfo{person}{Shizuo Tian}, \bibinfo{person}{Junda Zhang}, \bibinfo{person}{Jiahao Li}, \bibinfo{person}{Yunxin Li}, \bibinfo{person}{Shijue Huang}, {et~al\mbox{.}}} \bibinfo{year}{2025}\natexlab{}.
\newblock \showarticletitle{UI-TARS: Pioneering Automated GUI Interaction with Native Agents}.
\newblock \bibinfo{journal}{\emph{arXiv preprint arXiv:2501.12326}} (\bibinfo{year}{2025}).
\newblock


\bibitem[Radford et~al\mbox{.}(2021)]%
        {radford2021learningclip}
\bibfield{author}{\bibinfo{person}{Alec Radford}, \bibinfo{person}{Jong~Wook Kim}, \bibinfo{person}{Chris Hallacy}, \bibinfo{person}{Aditya Ramesh}, \bibinfo{person}{Gabriel Goh}, \bibinfo{person}{Sandhini Agarwal}, \bibinfo{person}{Girish Sastry}, \bibinfo{person}{Amanda Askell}, \bibinfo{person}{Pamela Mishkin}, \bibinfo{person}{Jack Clark}, {et~al\mbox{.}}} \bibinfo{year}{2021}\natexlab{}.
\newblock \showarticletitle{Learning transferable visual models from natural language supervision}. In \bibinfo{booktitle}{\emph{International conference on machine learning}}. PmLR, \bibinfo{pages}{8748--8763}.
\newblock


\bibitem[{Schneider Electric}({[n.\,d.]})]%
        {schneider_ecostruxure_it}
\bibfield{author}{\bibinfo{person}{{Schneider Electric}}.} \bibinfo{year}{[n.\,d.]}\natexlab{}.
\newblock \bibinfo{title}{EcoStruxure IT}.
\newblock \bibinfo{howpublished}{\url{https://www.se.com/en/work/solutions/for-business/data-centers-and-networks/dcim-software/what-is-ecostruxure-it.jsp}}.
\newblock


\bibitem[Singh et~al\mbox{.}(2025)]%
        {singh2025trishul}
\bibfield{author}{\bibinfo{person}{Kunal Singh}, \bibinfo{person}{Shreyas Singh}, {and} \bibinfo{person}{Mukund Khanna}.} \bibinfo{year}{2025}\natexlab{}.
\newblock \showarticletitle{TRISHUL: Towards Region Identification and Screen Hierarchy Understanding for Large VLM based GUI Agents}. In \bibinfo{booktitle}{\emph{Proceedings of the Computer Vision and Pattern Recognition Conference}}. \bibinfo{pages}{170--179}.
\newblock


\bibitem[Sobczak and Ziora(2021)]%
        {en14165191RPA4}
\bibfield{author}{\bibinfo{person}{Andrzej Sobczak} {and} \bibinfo{person}{Leszek Ziora}.} \bibinfo{year}{2021}\natexlab{}.
\newblock \showarticletitle{The Use of Robotic Process Automation (RPA) as an Element of Smart City Implementation: A Case Study of Electricity Billing Document Management at Bydgoszcz City Hall}.
\newblock \bibinfo{journal}{\emph{Energies}} \bibinfo{volume}{14}, \bibinfo{number}{16} (\bibinfo{year}{2021}).
\newblock
\showISSN{1996-1073}
\urldef\tempurl%
\url{https://doi.org/10.3390/en14165191}
\showDOI{\tempurl}


\bibitem[Syed et~al\mbox{.}(2020)]%
        {syed2020roboticRPA6}
\bibfield{author}{\bibinfo{person}{Rehan Syed}, \bibinfo{person}{Suriadi Suriadi}, \bibinfo{person}{Michael Adams}, \bibinfo{person}{Wasana Bandara}, \bibinfo{person}{Sander~JJ Leemans}, \bibinfo{person}{Chun Ouyang}, \bibinfo{person}{Arthur~HM Ter~Hofstede}, \bibinfo{person}{Inge Van De~Weerd}, \bibinfo{person}{Moe~Thandar Wynn}, {and} \bibinfo{person}{Hajo~A Reijers}.} \bibinfo{year}{2020}\natexlab{}.
\newblock \showarticletitle{Robotic process automation: contemporary themes and challenges}.
\newblock \bibinfo{journal}{\emph{Computers in industry}}  \bibinfo{volume}{115} (\bibinfo{year}{2020}), \bibinfo{pages}{103162}.
\newblock


\bibitem[Tang et~al\mbox{.}(2025)]%
        {tang2025guigrounding}
\bibfield{author}{\bibinfo{person}{Fei Tang}, \bibinfo{person}{Zhangxuan Gu}, \bibinfo{person}{Zhengxi Lu}, \bibinfo{person}{Xuyang Liu}, \bibinfo{person}{Shuheng Shen}, \bibinfo{person}{Changhua Meng}, \bibinfo{person}{Wen Wang}, \bibinfo{person}{Wenqi Zhang}, \bibinfo{person}{Yongliang Shen}, \bibinfo{person}{Weiming Lu}, {et~al\mbox{.}}} \bibinfo{year}{2025}\natexlab{}.
\newblock \showarticletitle{GUI-G$^2$: Gaussian Reward Modeling for GUI Grounding}.
\newblock \bibinfo{journal}{\emph{arXiv preprint arXiv:2507.15846}} (\bibinfo{year}{2025}).
\newblock


\bibitem[{The Green Grid}(2012)]%
        {greengrid_dcim}
\bibfield{author}{\bibinfo{person}{{The Green Grid}}.} \bibinfo{year}{2012}\natexlab{}.
\newblock \bibinfo{booktitle}{\emph{Data Center Automation with a DCIM System}}.
\newblock \bibinfo{type}{{T}echnical {R}eport} White Paper \#79. \bibinfo{institution}{The Green Grid}.
\newblock
\urldef\tempurl%
\url{https://www.thegreengrid.org/en/resources/library-and-tools/492-WP%2379---Data-Center-Automation-with-a-DCIM-System}
\showURL{%
\tempurl}


\bibitem[Wu et~al\mbox{.}(2025b)]%
        {wu-etal-2025-webwalker}
\bibfield{author}{\bibinfo{person}{Jialong Wu}, \bibinfo{person}{Wenbiao Yin}, \bibinfo{person}{Yong Jiang}, \bibinfo{person}{Zhenglin Wang}, \bibinfo{person}{Zekun Xi}, \bibinfo{person}{Runnan Fang}, \bibinfo{person}{Linhai Zhang}, \bibinfo{person}{Yulan He}, \bibinfo{person}{Deyu Zhou}, \bibinfo{person}{Pengjun Xie}, {and} \bibinfo{person}{Fei Huang}.} \bibinfo{year}{2025}\natexlab{b}.
\newblock \showarticletitle{{W}eb{W}alker: Benchmarking {LLM}s in Web Traversal}. In \bibinfo{booktitle}{\emph{Proceedings of the 63rd Annual Meeting of the Association for Computational Linguistics (Volume 1: Long Papers)}}, \bibfield{editor}{\bibinfo{person}{Wanxiang Che}, \bibinfo{person}{Joyce Nabende}, \bibinfo{person}{Ekaterina Shutova}, {and} \bibinfo{person}{Mohammad~Taher Pilehvar}} (Eds.). \bibinfo{publisher}{Association for Computational Linguistics}, \bibinfo{address}{Vienna, Austria}, \bibinfo{pages}{10290--10305}.
\newblock
\showISBNx{979-8-89176-251-0}
\urldef\tempurl%
\url{https://aclanthology.org/2025.acl-long.508/}
\showURL{%
\tempurl}


\bibitem[Wu et~al\mbox{.}(2025a)]%
        {wu2025backtrackagent}
\bibfield{author}{\bibinfo{person}{Qinzhuo Wu}, \bibinfo{person}{Pengzhi Gao}, \bibinfo{person}{Wei Liu}, {and} \bibinfo{person}{Jian Luan}.} \bibinfo{year}{2025}\natexlab{a}.
\newblock \showarticletitle{BacktrackAgent: Enhancing GUI Agent with Error Detection and Backtracking Mechanism}.
\newblock \bibinfo{journal}{\emph{arXiv preprint arXiv:2505.20660}} (\bibinfo{year}{2025}).
\newblock


\bibitem[Xie et~al\mbox{.}(2024)]%
        {xie2024osworld}
\bibfield{author}{\bibinfo{person}{Tianbao Xie}, \bibinfo{person}{Danyang Zhang}, \bibinfo{person}{Jixuan Chen}, \bibinfo{person}{Xiaochuan Li}, \bibinfo{person}{Siheng Zhao}, \bibinfo{person}{Ruisheng Cao}, \bibinfo{person}{Toh~J Hua}, \bibinfo{person}{Zhoujun Cheng}, \bibinfo{person}{Dongchan Shin}, \bibinfo{person}{Fangyu Lei}, {et~al\mbox{.}}} \bibinfo{year}{2024}\natexlab{}.
\newblock \showarticletitle{Osworld: Benchmarking multimodal agents for open-ended tasks in real computer environments}.
\newblock \bibinfo{journal}{\emph{Advances in Neural Information Processing Systems}}  \bibinfo{volume}{37} (\bibinfo{year}{2024}), \bibinfo{pages}{52040--52094}.
\newblock


\bibitem[Yadav and Paul(2021)]%
        {YADAV2021100433MCI4}
\bibfield{author}{\bibinfo{person}{Geeta Yadav} {and} \bibinfo{person}{Kolin Paul}.} \bibinfo{year}{2021}\natexlab{}.
\newblock \showarticletitle{Architecture and security of SCADA systems: A review}.
\newblock \bibinfo{journal}{\emph{International Journal of Critical Infrastructure Protection}}  \bibinfo{volume}{34} (\bibinfo{year}{2021}), \bibinfo{pages}{100433}.
\newblock
\showISSN{1874-5482}
\urldef\tempurl%
\url{https://doi.org/10.1016/j.ijcip.2021.100433}
\showDOI{\tempurl}


\bibitem[Yamamoto et~al\mbox{.}(2020)]%
        {en13092342RPA1}
\bibfield{author}{\bibinfo{person}{Toru Yamamoto}, \bibinfo{person}{Hirofumi Hayama}, \bibinfo{person}{Takao Hayashi}, {and} \bibinfo{person}{Taro Mori}.} \bibinfo{year}{2020}\natexlab{}.
\newblock \showarticletitle{Automatic Energy-Saving Operations System Using Robotic Process Automation}.
\newblock \bibinfo{journal}{\emph{Energies}} \bibinfo{volume}{13}, \bibinfo{number}{9} (\bibinfo{year}{2020}).
\newblock
\showISSN{1996-1073}
\urldef\tempurl%
\url{https://doi.org/10.3390/en13092342}
\showDOI{\tempurl}


\bibitem[Yang et~al\mbox{.}(2024)]%
        {10.1145/3699518knowdistsurvey}
\bibfield{author}{\bibinfo{person}{Chuanpeng Yang}, \bibinfo{person}{Yao Zhu}, \bibinfo{person}{Wang Lu}, \bibinfo{person}{Yidong Wang}, \bibinfo{person}{Qian Chen}, \bibinfo{person}{Chenlong Gao}, \bibinfo{person}{Bingjie Yan}, {and} \bibinfo{person}{Yiqiang Chen}.} \bibinfo{year}{2024}\natexlab{}.
\newblock \showarticletitle{Survey on Knowledge Distillation for Large Language Models: Methods, Evaluation, and Application}.
\newblock \bibinfo{journal}{\emph{ACM Trans. Intell. Syst. Technol.}} (\bibinfo{date}{Oct.} \bibinfo{year}{2024}).
\newblock
\showISSN{2157-6904}
\urldef\tempurl%
\url{https://doi.org/10.1145/3699518}
\showDOI{\tempurl}


\bibitem[Yigit et~al\mbox{.}(2025)]%
        {s25061666MCI5}
\bibfield{author}{\bibinfo{person}{Yagmur Yigit}, \bibinfo{person}{Mohamed~Amine Ferrag}, \bibinfo{person}{Mohamed~C. Ghanem}, \bibinfo{person}{Iqbal~H. Sarker}, \bibinfo{person}{Leandros~A. Maglaras}, \bibinfo{person}{Christos Chrysoulas}, \bibinfo{person}{Naghmeh Moradpoor}, \bibinfo{person}{Norbert Tihanyi}, {and} \bibinfo{person}{Helge Janicke}.} \bibinfo{year}{2025}\natexlab{}.
\newblock \showarticletitle{Generative AI and LLMs for Critical Infrastructure Protection: Evaluation Benchmarks, Agentic AI, Challenges, and Opportunities}.
\newblock \bibinfo{journal}{\emph{Sensors}} \bibinfo{volume}{25}, \bibinfo{number}{6} (\bibinfo{year}{2025}).
\newblock
\showISSN{1424-8220}
\urldef\tempurl%
\url{https://doi.org/10.3390/s25061666}
\showDOI{\tempurl}


\bibitem[Zangana et~al\mbox{.}(2025)]%
        {zangana2025leveragingMCI6}
\bibfield{author}{\bibinfo{person}{Hewa Zangana}, \bibinfo{person}{Natheer~Yaseen Ali}, \bibinfo{person}{Sameer Mohammed~Salih Bazeed}, {and} \bibinfo{person}{Dilovan~Taha Abdullah}.} \bibinfo{year}{2025}\natexlab{}.
\newblock \showarticletitle{Leveraging Automation and Traceability in Managing Changes to Mission-Critical Computer Systems}.
\newblock \bibinfo{journal}{\emph{Indonesian Journal of Education and Social Sciences}} \bibinfo{volume}{4}, \bibinfo{number}{1} (\bibinfo{year}{2025}), \bibinfo{pages}{176--189}.
\newblock


\bibitem[Zhang et~al\mbox{.}(2025)]%
        {zhang2025enhancing}
\bibfield{author}{\bibinfo{person}{Zhisong Zhang}, \bibinfo{person}{Tianqing Fang}, \bibinfo{person}{Kaixin Ma}, \bibinfo{person}{Wenhao Yu}, \bibinfo{person}{Hongming Zhang}, \bibinfo{person}{Haitao Mi}, {and} \bibinfo{person}{Dong Yu}.} \bibinfo{year}{2025}\natexlab{}.
\newblock \showarticletitle{Enhancing web agents with explicit rollback mechanisms}.
\newblock \bibinfo{journal}{\emph{arXiv preprint arXiv:2504.11788}} (\bibinfo{year}{2025}).
\newblock


\bibitem[Zhao et~al\mbox{.}(2025)]%
        {zhao2025worldguiinteractivebenchmarkdesktop}
\bibfield{author}{\bibinfo{person}{Henry~Hengyuan Zhao}, \bibinfo{person}{Kaiming Yang}, \bibinfo{person}{Wendi Yu}, \bibinfo{person}{Difei Gao}, {and} \bibinfo{person}{Mike~Zheng Shou}.} \bibinfo{year}{2025}\natexlab{}.
\newblock \bibinfo{title}{WorldGUI: An Interactive Benchmark for Desktop GUI Automation from Any Starting Point}.
\newblock
\newblock
\showeprint[arxiv]{2502.08047}~[cs.AI]
\urldef\tempurl%
\url{https://arxiv.org/abs/2502.08047}
\showURL{%
\tempurl}


\bibitem[Zheng et~al\mbox{.}(2025)]%
        {zheng2025vemenvironmentfreeexplorationtraining}
\bibfield{author}{\bibinfo{person}{Jiani Zheng}, \bibinfo{person}{Lu Wang}, \bibinfo{person}{Fangkai Yang}, \bibinfo{person}{Chaoyun Zhang}, \bibinfo{person}{Lingrui Mei}, \bibinfo{person}{Wenjie Yin}, \bibinfo{person}{Qingwei Lin}, \bibinfo{person}{Dongmei Zhang}, \bibinfo{person}{Saravan Rajmohan}, {and} \bibinfo{person}{Qi Zhang}.} \bibinfo{year}{2025}\natexlab{}.
\newblock \bibinfo{title}{VEM: Environment-Free Exploration for Training GUI Agent with Value Environment Model}.
\newblock
\newblock
\showeprint[arxiv]{2502.18906}~[cs.LG]
\urldef\tempurl%
\url{https://arxiv.org/abs/2502.18906}
\showURL{%
\tempurl}


\end{thebibliography}

\end{document}